\documentclass[acmsmall]{acmart}

\AtBeginDocument{%
  }

\setcopyright{cc}
\setcctype{by}
\acmYear{2025} 
\acmDOI{10.1145/3769757}

\acmJournal{PACMMOD}
\acmVolume{3} 
\acmNumber{6 (SIGMOD)} 
\acmArticle{292} 
\acmMonth{12} 

\acmPrice{}

\usepackage{balance}
\usepackage{subcaption}
\usepackage{tabularx}
\usepackage{xspace}
\usepackage{colortbl}
\usepackage{wrapfig}
\usepackage{tikz}
\usepackage{algorithm}
\usepackage[noend]{algorithmic}
\usepackage{listings}
\usepackage{enumitem}
\usepackage{amsmath}
\usepackage{algorithm}
\usepackage{algorithmic}
\usepackage{layouts}
\usepackage{pifont}
\usepackage{multirow}
\usepackage{thmtools}
\usepackage{subcaption}
\usepackage{tabularx}
\usepackage{booktabs}
\usepackage{tabularx}
\usepackage{lscape}

\usepackage{pgfplots}
\usepackage{pgfplotstable}
\usepackage{filecontents}
\usepgfplotslibrary{statistics}
\pgfplotsset{compat=1.17}

\usepackage[
    skins,
    breakable
]{tcolorbox}

\usepackage{lipsum}
\usepackage{url}

\newcommand{\framework}{\textsc{Aegis}\xspace}
\newcommand{\Dom}{\texttt{Dom}}
\newcommand{\Dist}{\texttt{Dist}}

\newtheoremstyle{defstyle}
{0.5pt plus 0pt minus 0pt}
{0pt plus 0pt minus 0pt}
{\it}
{\parindent}
{\sc}
{.}
{0.5em}
{}

{
\theoremstyle{defstyle}
\newtheorem{lemma}{Lemma}

\newtheorem*{problem*}{Problem Statement}
\newtheorem{example}{Example}
}

\usepackage{array}
\usepackage{mathrsfs}
\usepackage{amsxtra}

\newcommand{\eat}[1]{} % TO MAKE LARGE BLOCKS OF TEXT INVISIBLE

\newcommand{\sz}[1]{\lvert#1\rvert}   % cardinality of set
\newcommand{\td}[2]{\if*#1\else^{#1}\fi\if*#2\else_{#2}\fi} % ^?_?, *=ignore

\DeclareMathOperator*{\argmin}{argmin} % amsmath package required
\newcommand{\sset}[1]{\left\{\,#1\,\right\}} % { ? }, automatic brackets
\newcommand\join\Join % normal join

\DeclareSymbolFont{txsymbolsC}{U}{txsyc}{m}{n}
\SetSymbolFont{txsymbolsC}{bold}{U}{txsyc}{bx}{n}
\DeclareFontSubstitution{U}{txsyc}{m}{n}
\DeclareMathSymbol{\ljoin}{\mathrel}{txsymbolsC}{88}
\DeclareMathSymbol{\rjoin}{\mathrel}{txsymbolsC}{89}
\newcommand\sel\sigma
\newcommand\proj\pi
\newcommand\cross\times

\newcommand\LRA\Leftrightarrow

\newsavebox\setminusbox
\newlength\setminuslen
\newcolumntype{C}{>{$\displaystyle}c<{$}} % centered math
\newcolumntype{L}{>{$\displaystyle}l<{$}} % left math
\newcolumntype{R}{>{$\displaystyle}r<{$}} % right math

\newcommand{\B}[3]{B\if*#1\else_{#1}\fi(#2,#3)} % (incomplete) beta function
\newcommand{\I}[3]{I\if*#1\else_{#1}\fi(#2,#3)} % regularized incomplete beta function
\makeatletter
\def\imod#1{\allowbreak\mkern10mu({\operator@font mod}\,\,#1)}
\makeatother

\newlength\hspaceoflen

\newcommand\vect[1]{{\boldsymbol{#1}}}
\newcommand\va{\vect{a}}
\newcommand\vb{\vect{b}}
\newcommand\vc{\vect{c}}
\newcommand\vd{\vect{d}}
\newcommand\ve{\vect{e}}
\newcommand\vf{\vect{f}}
\newcommand\vg{\vect{g}}
\newcommand\vh{\vect{h}}
\newcommand\vi{\vect{i}}
\newcommand\vj{\vect{j}}
\newcommand\vk{\vect{k}}
\newcommand\vl{\vect{l}}
\newcommand\vm{\vect{m}}
\newcommand\vn{\vect{n}}
\newcommand\vo{\vect{o}}
\newcommand\vp{\vect{p}}
\newcommand\vq{\vect{q}}
\newcommand\vr{\vect{r}}
\newcommand\vs{\vect{s}}
\newcommand\vt{\vect{t}}
\newcommand\vu{\vect{u}}
\newcommand\vw{\vect{w}}
\newcommand\vx{\vect{x}}
\newcommand\vy{\vect{y}}
\newcommand\vz{\vect{z}}

\newcommand\mA{\vect{A}}
\newcommand\mB{\vect{B}}
\newcommand\mC{\vect{C}}
\newcommand\mD{\vect{D}}
\newcommand\mE{\vect{E}}
\newcommand\mF{\vect{F}}
\newcommand\mG{\vect{G}}
\newcommand\mH{\vect{H}}
\newcommand\mI{\vect{I}}
\newcommand\mJ{\vect{J}}
\newcommand\mK{\vect{K}}
\newcommand\mL{\vect{L}}
\newcommand\mM{\vect{M}}
\newcommand\mN{\vect{N}}
\newcommand\mO{\vect{O}}
\newcommand\mP{\vect{P}}
\newcommand\mQ{\vect{Q}}
\newcommand\mR{\vect{R}}
\newcommand\mS{\vect{S}}
\newcommand\mT{\vect{T}}
\newcommand\mU{\vect{U}}
\newcommand\mV{\vect{V}}
\newcommand\mW{\vect{W}}
\newcommand\mX{\vect{X}}
\newcommand\mY{\vect{Y}}
\newcommand\mZ{\vect{Z}}

\newcommand\bF{\mathbb{F}}

\newcommand\bI{\mathbb{I}}

\newcommand\bT{\mathbb{T}}

\newcommand\bX{\mathbb{X}}
\newcommand\bY{\mathbb{Y}}
\DeclareMathAlphabet{\mathcal}{OMS}{cmsy}{m}{n}

\newcommand\cC{\mathcal{C}}

\newcommand\cM{\mathcal{M}}

\accentedsymbol\Ahat{{\hat A}}
\accentedsymbol\Bhat{{\hat B}}
\accentedsymbol\Chat{{\hat C}}
\accentedsymbol\Dhat{{\hat D}}
\accentedsymbol\Ehat{{\hat E}}
\accentedsymbol\Fhat{{\hat F}}
\accentedsymbol\Ghat{{\hat G}}
\accentedsymbol\Hhat{{\hat H}}
\accentedsymbol\Ihat{{\hat I}}
\accentedsymbol\Jhat{{\hat J}}
\accentedsymbol\Khat{{\hat K}}
\accentedsymbol\Lhat{{\hat L}}
\accentedsymbol\Mhat{{\hat M}}
\accentedsymbol\Nhat{{\hat N}}
\accentedsymbol\Ohat{{\hat O}}
\accentedsymbol\Phat{{\hat P}}
\accentedsymbol\Qhat{{\hat Q}}
\accentedsymbol\Rhat{{\hat R}}
\accentedsymbol\Shat{{\hat S}}
\accentedsymbol\That{{\hat T}}
\accentedsymbol\Uhat{{\hat U}}
\accentedsymbol\Vhat{{\hat V}}
\accentedsymbol\What{{\hat W}}
\accentedsymbol\Xhat{{\hat X}}
\accentedsymbol\Yhat{{\hat Y}}
\accentedsymbol\Zhat{{\hat Z}}

\accentedsymbol\ahat{{\hat a}}
\accentedsymbol\bhat{{\hat b}}
\accentedsymbol\chat{{\hat c}}
\accentedsymbol\dhat{{\hat d}}
\accentedsymbol\ehat{{\hat e}}
\accentedsymbol\fhat{{\hat f}}
\accentedsymbol\ghat{{\hat g}}
\accentedsymbol\hhat{{\hat h}}
\accentedsymbol\ihat{{\hat i}}
\accentedsymbol\jhat{{\hat j}}
\accentedsymbol\khat{{\hat k}}
\accentedsymbol\lhat{{\hat l}}
\accentedsymbol\mhat{{\hat m}}
\accentedsymbol\nhat{{\hat n}}
\accentedsymbol\ohat{{\hat o}}
\accentedsymbol\phat{{\hat p}}
\accentedsymbol\qhat{{\hat q}}
\accentedsymbol\rhat{{\hat r}}
\accentedsymbol\shat{{\hat s}}
\accentedsymbol\that{{\hat t}}
\accentedsymbol\uhat{{\hat u}}
\accentedsymbol\vhat{{\hat v}}
\accentedsymbol\what{{\hat w}}
\accentedsymbol\xhat{{\hat x}}
\accentedsymbol\yhat{{\hat y}}
\accentedsymbol\zhat{{\hat z}}

\accentedsymbol\rhohat{{\hat\rho}}

\accentedsymbol\Abar{{\bar A}}
\accentedsymbol\Bbar{{\bar B}}
\accentedsymbol\Cbar{{\bar C}}
\accentedsymbol\Dbar{{\bar D}}
\accentedsymbol\Ebar{{\bar E}}
\accentedsymbol\Fbar{{\bar F}}
\accentedsymbol\Gbar{{\bar G}}
\accentedsymbol\Hbar{{\bar H}}
\accentedsymbol\Ibar{{\bar I}}
\accentedsymbol\Jbar{{\bar J}}
\accentedsymbol\Kbar{{\bar K}}
\accentedsymbol\Lbar{{\bar L}}
\accentedsymbol\Mbar{{\bar M}}
\accentedsymbol\Nbar{{\bar N}}
\accentedsymbol\Obar{{\bar O}}
\accentedsymbol\Pbar{{\bar P}}
\accentedsymbol\Qbar{{\bar Q}}
\accentedsymbol\Rbar{{\bar R}}
\accentedsymbol\Sbar{{\bar S}}
\accentedsymbol\Tbar{{\bar T}}
\accentedsymbol\Ubar{{\bar U}}
\accentedsymbol\Vbar{{\bar V}}
\accentedsymbol\Wbar{{\bar W}}
\accentedsymbol\Xbar{{\bar X}}
\accentedsymbol\Ybar{{\bar Y}}
\accentedsymbol\abar{{\bar a}}
\accentedsymbol\bbar{{\bar b}}
\accentedsymbol\cbar{{\bar c}}
\accentedsymbol\dbar{{\bar d}}
\accentedsymbol\ebar{{\bar e}}
\accentedsymbol\fbar{{\bar f}}
\accentedsymbol\gbar{{\bar g}}
\makeatletter
\@ifundefined{hbar}{}{
        \let\hbar\@undefined
}
\makeatother
\accentedsymbol\hbar{{\bar h}}
\accentedsymbol\ibar{{\bar i}}
\accentedsymbol\jbar{{\bar j}}
\accentedsymbol\kbar{{\bar k}}
\accentedsymbol\lbar{{\bar l}}
\accentedsymbol\mbar{{\bar m}}
\accentedsymbol\nbar{{\bar n}}
\makeatletter
\@ifundefined{obar}{}{
        \let\obar\@undefined
}
\makeatother
\accentedsymbol{\obar}{{\bar o}}
\accentedsymbol\pbar{{\bar p}}
\accentedsymbol\qbar{{\bar q}}
\accentedsymbol\rbar{{\bar r}}
\accentedsymbol\sbar{{\bar s}}
\accentedsymbol\tbar{{\bar t}}
\accentedsymbol\ubar{{\bar u}}
\accentedsymbol\vbar{{\bar v}}
\accentedsymbol\wbar{{\bar w}}
\accentedsymbol\xbar{{\bar x}}
\accentedsymbol\ybar{{\bar y}}
\accentedsymbol\zbar{{\bar z}}

\renewcommand{\epsilon}{\varepsilon}

\accentedsymbol\mAhat{{\hat\mA}}
\accentedsymbol\mBhat{{\hat\mB}}
\accentedsymbol\mChat{{\hat\mC}}
\accentedsymbol\mDhat{{\hat\mD}}
\accentedsymbol\mEhat{{\hat\mE}}
\accentedsymbol\mFhat{{\hat\mF}}
\accentedsymbol\mGhat{{\hat\mG}}
\accentedsymbol\mHhat{{\hat\mH}}
\accentedsymbol\mIhat{{\hat\mI}}
\accentedsymbol\mJhat{{\hat\mJ}}
\accentedsymbol\mKhat{{\hat\mK}}
\accentedsymbol\mLhat{{\hat\mL}}
\accentedsymbol\mMhat{{\hat\mM}}
\accentedsymbol\mNhat{{\hat\mN}}
\accentedsymbol\mOhat{{\hat\mO}}
\accentedsymbol\mPhat{{\hat\mP}}
\accentedsymbol\mQhat{{\hat\mQ}}
\accentedsymbol\mRhat{{\hat\mR}}
\accentedsymbol\mShat{{\hat\mS}}
\accentedsymbol\mThat{{\hat\mT}}
\accentedsymbol\mUhat{{\hat\mU}}
\accentedsymbol\mVhat{{\hat\mV}}
\accentedsymbol\mWhat{{\hat\mW}}
\accentedsymbol\mXhat{{\hat\mX}}
\accentedsymbol\mYhat{{\hat\mY}}
\accentedsymbol\mZhat{{\hat\mZ}}

\accentedsymbol\vahat{{\hat\va}}
\accentedsymbol\vbhat{{\hat\vb}}
\accentedsymbol\vchat{{\hat\vc}}
\accentedsymbol\vdhat{{\hat\vd}}
\accentedsymbol\vehat{{\hat\ve}}
\accentedsymbol\vfhat{{\hat\vf}}
\accentedsymbol\vghat{{\hat\vg}}
\accentedsymbol\vhhat{{\hat\vh}}
\accentedsymbol\vihat{{\hat\vi}}
\accentedsymbol\vjhat{{\hat\vj}}
\accentedsymbol\vkhat{{\hat\vk}}
\accentedsymbol\vlhat{{\hat\vl}}
\accentedsymbol\vmhat{{\hat\vm}}
\accentedsymbol\vnhat{{\hat\vn}}
\accentedsymbol\vohat{{\hat\vo}}
\accentedsymbol\vphat{{\hat\vp}}
\accentedsymbol\vqhat{{\hat\vq}}
\accentedsymbol\vrhat{{\hat\vr}}
\accentedsymbol\vshat{{\hat\vs}}
\accentedsymbol\vthat{{\hat\vt}}
\accentedsymbol\vuhat{{\hat\vu}}
\accentedsymbol\vwhat{{\hat\vw}}
\accentedsymbol\vxhat{{\hat\vx}}
\accentedsymbol\vyhat{{\hat\vy}}
\accentedsymbol\vzhat{{\hat\vz}}

\begin{document}

\title{\framework: A Correlation-Based Data Masking Advisor for Data-Sharing Ecosystems}

\author{Omar Islam Laskar}
\affiliation{%
  \institution{Indian Institute of Technology Delhi}
  \city{New Delhi}
  \country{India}
  }
\email{omar.laskar.cs522@cse.iitd.ac.in}

\author{Fatemeh Ramezani Khozestani}
\affiliation{%
  \institution{New Jersey Institute of Technology}
  \city{New Jersey}
  \country{USA}
  }
\email{fr46@njit.edu}

\author{Ishika Nankani}
\affiliation{%
  \institution{Indian Institute of Technology Delhi}
  \city{New Delhi}
  \country{India}
  }
\email{ishika.nankani.cs122@cse.iitd.ac.in}

\author{Sohrab Namazi Nia}
\affiliation{%
  \institution{New Jersey Institute of Technology}
  \city{New Jersey}
  \country{USA}
  }
\email{sn773@njit.edu}

\author{Senjuti Basu Roy}
\affiliation{%
  \institution{New Jersey Institute of Technology}
  \city{New Jersey}
  \country{USA}
  }
\email{senjuti.basuroy@njit.edu}

\author{Kaustubh Beedkar}
\affiliation{%
  \institution{Indian Institute of Technology Delhi}
  \city{New Delhi}
  \country{India}
  }
\email{kbeedkar@cse.iitd.ac.in}

\renewcommand{\shortauthors}{Omar Islam Laskar et al.}
%% No italics, no superscripts
%% If needed use a foot or author note to identify equal contribution

\begin{abstract}
 Data-sharing ecosystems connect providers, consumers, and intermediaries to facilitate the exchange and use of data for a wide range of downstream tasks. In sensitive domains such as healthcare, privacy is enforced as a hard constraint---any shared data must satisfy a minimum privacy threshold. However, among all masking configurations that meet this requirement, the utility of the masked data can vary significantly, posing a key challenge: how to efficiently select the \emph{optimal} configuration that preserves maximum utility. This paper presents \framework{}, a middleware framework that selects optimal masking configurations for machine learning datasets with features and class labels. \framework{} incorporates a utility optimizer that minimizes \emph{predictive utility deviation}---quantifying shifts in feature-label correlations due to masking. Our framework leverages limited data summaries (such as 1D histograms) or none to estimate the feature-label joint distribution, making it suitable for scenarios where raw data is inaccessible due to privacy restrictions. To achieve this, we propose a joint distribution estimator based on iterative proportional fitting, which allows supporting various feature-label correlation quantification methods such as mutual information, chi-square, or g3. Our experimental evaluation of real-world datasets shows that \framework{} identifies optimal masking configurations over an order of magnitude faster, while the resulting masked datasets achieve predictive performance on downstream ML tasks on par with baseline approaches and complements privacy anonymization data masking techniques.
 \end{abstract}

\begin{CCSXML}
<ccs2012>
   <concept>
       <concept_id>10002951.10002952</concept_id>
       <concept_desc>Information systems~Data management systems</concept_desc>
       <concept_significance>500</concept_significance>
       </concept>
   <concept>
       <concept_id>10002978.10003018.10003019</concept_id>
       <concept_desc>Security and privacy~Data anonymization and sanitization</concept_desc>
       <concept_significance>500</concept_significance>
       </concept>
 </ccs2012>
\end{CCSXML}

\ccsdesc[500]{Information systems~Data management systems}
\ccsdesc[500]{Security and privacy~Data anonymization and sanitization}

\keywords{Data Markets, Data Sharing Ecosystems, Data Masking, Compliant Data Management}

\received{April 2025}
\received[revised]{July 2025}
\received[accepted]{August 2025}

\maketitle

%!TEX root=./main.tex

\section{Introduction} % (fold)
\label{sec:introduction}
The increasing adoption of data-driven decision-making and machine learning (ML) in finance, healthcare, and public policy has made data-sharing ecosystems an integral component of modern information infrastructure. These ecosystems bring together data providers, consumers, and intermediaries---such as AWS Data Exchange~\cite{aws}, Snowflake Data Marketplace~\cite{snowflake}, OpenMined~\cite{openmined2023}, DataSHIELD~\cite{datashield2023}, GAIA-X~\cite{gaiax2023}) and open data portals~\cite{iudx}---to facilitate the access, exchange, and use of data for research, innovation, and economic development across both public and private sectors.

As data-sharing practices become more widespread, concerns around data privacy and proprietorship have led to the need for strict compliance with data regulations like GDPR~\cite{gdpr2016}, CCPA~\cite{ccpa2020}, HIPAA~\cite{hipaa1996} among others. As a result, data providers often anonymize their datasets by masking sensitive attributes using masking functions like generalization, which replaces specific values with broader categories; suppression, which redacts sensitive details; and perturbation, which introduces controlled noise. While anonymization is essential for regulatory compliance, it often transforms key properties of the data---posing challenges for downstream ML tasks, where preserving statistical structure and predictive performance is crucial. Moreover, for any given dataset, there are often multiple \emph{masking configurations}, each prescribing and differing in how specific attribute values must be masked~\cite{mascara}.

For example, consider the following scenario. In a real-world data marketplace for healthcare, hospitals and clinics share patient-level Electronic Health Records (EHRs) to third parties, such as insurance companies. In such settings (e.g., HIPAA in the U.S.) {\it privacy is binary}---the data is considered private and eligible for sharing if it satisfies a threshold, such as $k$-anonymity with a specific $k$-value~\cite{portability2012guidance,el2008protecting}; While a minimum $k$-value of 3 is sometimes suggested, a common recommendation in practice is to use $k=5$. However, some scenarios might require higher values, like $k=10$ or $15$, to reduce the risk of re-identification to an acceptable level is used.  When the specific privacy guarantee is not reached, the data is not private and cannot be shared. This binary treatment is crucial for compliance and auditability.
To satisfy the privacy requirement, the marketplace applies various masking configurations to the data.

In the above scenario for instance, consider a dataset for predicting health based on features like age, weight, and zip code. A detailed running example based on this dataset is presented in Section~\ref{sec:preliminaries}. Masking configurations may involve generalizing age into ranges (e.g., 23 $\to$ 20--30), blurring weight by rounding to the nearest ten (e.g., 53.5 $\to$ 50, or generalizing zip codes into broader geographic regions (e.g., 12345 $\to$ 123$**$). Alternatively, they might suppress zip codes entirely or add noise to sensitive numerical values. The system only allows those masking configurations that surpass the $k$-anonymity threshold, are considered valid, and only those versions of the data can be shared. In general, while each of the masking strategies that meet the binary privacy rule is acceptable, they affect the feature-label distributions differently and, consequently, may lead to substantial variation in the accuracy of the trained model.

\begin{figure}[t]
\centering
\includegraphics[width=0.6\linewidth]{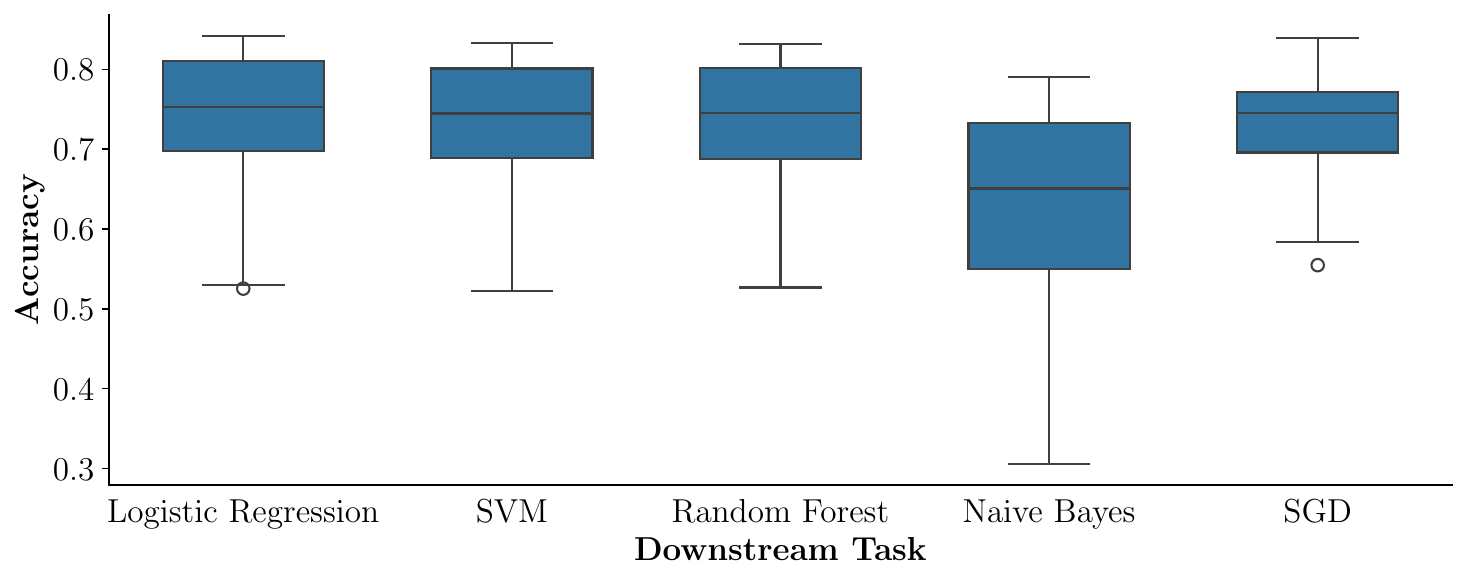}
\caption{Distribution of model accuracy across 50 masking configurations for Air Quality dataset~\cite{airquality}.}
\label{fig:accuracy_distribution}
\end{figure}

\begin{figure}[t]
% \setlength{\abovecaptionskip}{0mm}
% \setlength{\belowcaptionskip}{-5mm}
% \captionsetup[subfigure]{aboveskip=0pt,belowskip=2pt}
  \centering
  \begin{subfigure}{\linewidth}
    \centering
    \includegraphics[page=2]{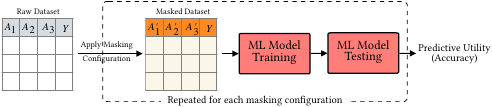}
    \caption{Exhaustive approach (traditional)}
    \label{fig:exhaustive_approach}
  \end{subfigure}

  \begin{subfigure}{\linewidth}
    \centering
    \includegraphics[page=3]{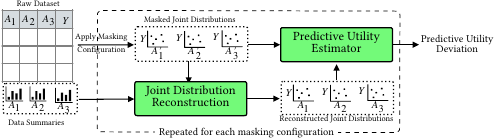}
    \caption{Our approach (proposed)}
    \label{fig:aegis_approach}
  \end{subfigure}
    \caption{Determining optimal masking configuration using (a)~traditional approach and (b)~our proposed approach.}
  \label{fig:comparison_overview}
\end{figure}

In the context of a data-sharing ecosystem, this creates a fundamental tension. Data providers often do not know the specific downstream task consumers intend to perform. They thus cannot select the masking configuration that will preserve their dataset's utility for a particular model or use case. For example, Figure~\ref{fig:accuracy_distribution} shows the results of an experiment (see Section~\ref{sec:evaluation} for details) that shows how a dataset's utility (based on model accuracy) changes across different masking configurations for a dataset depending on the downstream ML task. On the other hand, for data consumers, the state-of-the-art approach for finding the optimal masking configuration involves exhaustively applying each configuration, building a model, and evaluating its predictive utility (see Figure~\ref{fig:exhaustive_approach}). After testing all configurations, the one with the highest utility (e.g. highest accuracy) is selected. However, this exhaustive process is computationally expensive and impractical for large-scale real-world datasets and complex models.
This makes it difficult to efficiently find an effective masking configuration from a set of possible ones- one that preserves utility across a range of possible uses- both critical and nontrivial.

%!TEX root=./main.tex

\paragraph{Contributions.} 
In this paper, we present \framework{} (Figure~\ref{fig:aegis_approach}), a framework designed to efficiently selecting optimal masking configurations for machine learning datasets---without requiring exhaustive evaluation of the entire pipeline. Our work is an effort to
select masking configurations and not to create new ones that are
privacy aware.  Our work complements existing privacy preservation based masking techniques---\framework can incorporate any $k$-anonymity, differential privacy, structural anonymization, or masking operation to evaluate and select the configuration that best preserves predictive utility in a model-agnostic way. In that sense, we are orthogonal to existing work on
privacy preserving masking configuration generation related work.

At the core of \framework{} is its \emph{predictive utility estimator}, which estimates the predictive value of each feature with respect to the class label, even in scenarios where the feature, the label, or both are masked. This capability makes \framework{} well-suited as a middleware solution for data-sharing ecosystems where access to raw data is restricted. 

\framework{} builds upon the understanding that the predictive utility of a dataset is often evaluated through model-agnostic techniques, commonly referred to as correlation measures. A variety of methods exists for quantifying such relationships. Among the most widely used are: Mutual Information (MI), which captures the statistical dependency between variables~\cite{chen2021efficient, vergara2014review, pascoal2017theoretical, huang2007hybrid, battiti1994using, fleuret2004fast, kwak2002input}; the Chi-Square Test of Independence~\cite{cochran1952chi2}, which determines whether the two categorical variables are related by comparing observed and expected frequencies. More recently, \cite{le2020evaluating} has connected measures of approximate functional dependency, particularly the $g_3$ error~\cite{KIVINEN1995129}, to theoretical upper bounds on classification accuracy. Our predictive utility quantifier can accommodate any of these correlation measures as inputs and quantifies the impact of masking configurations on the dataset’s predictive utility. This allows masking configurations to be determined in a task-agnostic fashion. \framework{} is orthogonal to $k$-anonymity~\cite{sweeney2002k} and other privacy-preserving techniques~\cite{Li2007tcloseness, Machanavajjhala2006ldiversity}, allowing it to integrate seamlessly with existing privacy-preserving methods to maintain both privacy and predictive utility.

We propose the notion of {\em predictive utility deviation} inside {\em predictive utility estimator} in \framework{}. Given a set of predictors, a class label, a set of masking functions to be applied to the predictors, and a correlation measure, predictive utility deviation quantifies {\em the total change in correlation between the unmasked predictors and the class label compared to the masked predictors and the class label.} We then define the {\bf Optimal Predictive Utility Aware Masking Configuration Selection Problem}, as follows: Given a set of possible masking configurations to be applied over a set of predictors (attributes) and a class label, where the joint frequency distributions between the unmasked predictors and the class label are unknown (due to the sensitive nature of the predictors), select the masking configuration to be applied on the predictors, such that, the {\em predictive utility deviation is minimized}.

To minimize predictive utility deviation, the grand challenge is to be able to estimate the unmasked predictor from the known information and reconstruct the joint distribution of the predictor and the class label from there. Examples of known information could be 1D histograms of the predictor, e.g., for {\em Age},  for example, there exists $4$ individuals with age $10$, $12$ individuals $17$, $4$ with age $43$, and so on. Given the masked {\em Age}, it may be known that there exist 20 young and 80 old individuals in the masked Age, and the masking function dictates that {\em Young  $\rightarrow 10-45$, Old $\rightarrow > 45$}. From this, the reconstructed Age has to satisfy the following: there are $100$ individuals of {\em Age} between $10-80$, among them $4$ individuals are of age $10$, $12$ individuals of age $17$, $4$ of age $43$, and so on. Clearly, a prohibitively large number of possible distributions of {\em Age} satisfies all these constraints, and any of these could be a likely solution. Without additional information, we assume that any value of {\em Age} between $10-80$ is equally likely. We formalize this as an optimization problem, i.e., estimate the joint distribution of each predictor and the class label (i.e., {\em Age} and {\em Health}) such that the distribution of the predictor  (i.e., {\em Age}) is as uniform as possible within the given range (i.e., between $10-80$,). Still, the reconstructed joint distribution satisfies all constraints imposed on the marginals (e.g., available known information on {\em Age}). Once the joint distribution is estimated for each predictor-masking-function pair per configuration, the problem selects the one that minimizes the predictive utility deviation. We solve this problem by adapting Iterative Proportional Fitting (or IPF)~\cite{kruithof1937calculation,idel2016review}, an iterative algorithm designed to adjust statistics distribution.

We study two variants of the {\em Optimal Predictive Utility Aware Masking Configuration Selection Problem} : (a)~When 1D histograms of unmasked predictors are available. Additionally, masking functions to be applied on the predictors and the masked predictors are also known; (b)~When no distributional statistics of the unmasked predictors are available, but the masking functions to be applied on the predictors are known along with their masked values. We formalize both variants as optimization problems and demonstrate how IPF could be adapted to solve both. When more information is known about the predictors (case a), the joint estimation becomes more accurate.

We demonstrate the efficacy of the proposed framework using three real-world datasets considering three different correlation measures, namely, {\em Mutual Information, Chi-Squared Test, and $g_3$ error.} Our experiments demonstrate that \framework can determine an optimal masking configuration order of magnitude faster while achieving comparable predictive performance for various downstream ML tasks compared to baseline approaches.

Section~\ref{sec:preliminaries} presents data model and preliminaries, Section~\ref{sec:framework_overview} presents \framework{}. In Section~\ref{sec:reconstructing_join_distributions}, we present the algorithms inside \framework{}. Section~\ref{sec:evaluation} presents our experimental evaluation, we discuss related works in Section~\ref{sec:related_work}, and we conclude in Section~\ref{sec:conclusion}.

%!TEX root=./main.tex

%!TEX root=./main.tex

\begin{figure}[t]
% \setlength{\abovecaptionskip}{0mm}
% \setlength{\belowcaptionskip}{-5mm}
% \captionsetup[subfigure]{aboveskip=0pt,belowskip=2pt}
  \centering
  \small
  % Left table
  \begin{minipage}[b]{0.48\columnwidth}
  % \vspace{0pt}
    \centering
    \setlength\tabcolsep{2pt}
    \begin{tabular}{cccl}
      \toprule
      \textbf{Age} & \textbf{Weight} & \textbf{Zip} & \textbf{Health} \\
      \midrule
      10 & 30 & 21162 & \textbf{G}ood \\
      10 & 31 & 21168 & \textbf{G}ood \\
      43 & 63 & 22170 & \textbf{M}oderate \\
      65 & 71 & 23175 & \textbf{P}oor \\
      75 & 80 & 23173 & \textbf{V}ery \textbf{P}oor \\
      80 & 78 & 25165 & \textbf{V}ery \textbf{G}ood \\
      \bottomrule
    \end{tabular}
    \subcaption{Excerpt of a dataset}
    \label{tab:excerpt_dataset}
  \end{minipage}%
  \hfil
  % Right top & bottom tables
  \begin{minipage}[b]{0.48\columnwidth}
    \centering
    % \vspace{0pt}
    \setlength\tabcolsep{2pt}
    % Top: Age distribution
    \begin{tabular}{@{} l *{8}{c} @{}}
      \toprule
      \textbf{Age} & \textbf{10} & \textbf{17} & \textbf{43} & \textbf{55} & \textbf{60} & \textbf{65} & \textbf{75} & \textbf{80} \\
      \midrule
      \textbf{Count}
        &  4 & 12 &  4 & 30 & 20 & 10 & 10 & 10 \\
      \bottomrule
    \end{tabular}
    
    \vspace{0.8em}
    
    % Bottom: Health distribution
    \begin{tabular}{@{} l *{5}{c} @{}}
      \toprule
      \textbf{Health} & \textbf{VG} & \textbf{G} & \textbf{M} & \textbf{P} & \textbf{VP} \\
      \midrule
      \textbf{Count}
        & 14 & 16 & 30 & 25 & 15 \\
      \bottomrule
    \end{tabular}
    \subcaption{Marginal distributions}
    \label{tab:example_marginals}
  \end{minipage}
  \caption{(a) Example ML dataset with three attributes (Age, Weight, and Zip) and a label (Health); (b) marginal Age and Health distributions of 100 individuals.}
  \label{fig:example_dataset}
\end{figure}

\section{Preliminaries} % (fold)
\label{sec:preliminaries}

We introduce the data model we are considering and the related concepts. Table~\ref{tab:notations} summarizes the notations we commonly use throughout the paper.

\paragraph{Dataset}
We consider ML datasets, where a dataset is a collection of related observations comprising features/ attributes/predictors and a label organized in tabular format. More formally, denote by $D = \sset{(x_i, y_i)}_{i=1}^N$ a dataset with $N$ rows (data points). The i-th data point $x_i = (x_{i1}, x_{i2}, \dots, x_{im}) \in \bX^m)$ is a feature vector over $m$ attributes/predictors $(\mathcal{A}=\{A_1,A_2 \ldots A_m\})$. For the $i$-th data point, $y_i \in \bY$ is the label associated. 

\begin{example}\label{ex:dataset}
Figure~\ref{tab:excerpt_dataset} presents a sample dataset intended for machine learning tasks, where \texttt{age}, \texttt{weight}, and \texttt{zip} are attributes (predictors), and \texttt{health} serves as the label. Each feature contributes to predicting the health (class label).
\end{example}

\paragraph{Domain of an attribute}
 For an attribute $A$, $\Dom(A)$ denotes its domain, referring to all possible values that $A$ can take. The $j$-th domain value for $A$ is called $a^j$. Similarly, the domain of the label $\bY$ is $\Dom(\bY)$. For example, in Figure~\ref{tab:excerpt_dataset}, we have $\Dom$(Health)= \{Very Good, Good, Moderate, Poor, Very Poor\}.

\paragraph{Masking Functions}
In data-sharing ecosystems, datasets are often shared after anonymizing them by using masking functions. Moreover, an attribute (e.g., \texttt{Age}) can be masked differently using different masking functions (e.g., generalizing age to ranges into categories). For each attribute $A \in \mathcal{A}$, let $\mathcal{C}_A$ be the set of candidate masking functions applicable to $A$. When an attribute $A$ is masked using a masking function $M \in \mathcal{C}_A$, the resulting masked attribute is denoted by $M(A)$. More formally, a \emph{masking function}\footnote{In this paper, we consider scalar and value-based masking functions. In general, this is not a limitation. The techniques proposed here can be extended to include other masking functions.} $M$ is a function that takes as input the $i$-th domain value $a^i \in \Dom(A)$ (typically with other function parameters) and outputs a masked value $a^{i'}$, such that:
$ M(a^i) = a^{i'}, \quad \forall a^i \in \text{Dom}(A).$
The masked attribute $M(A)$ is then defined as the attribute whose domain consists of the values $\{M(a^i) \mid a^i \in \text{Dom}(A)\}$. 

\paragraph{Domain of a masked attribute}
Let $\Dom(M(A))$ denote the domain of attribute $A$ masked by the masking function $M$. Using our running example, we have $\Dom(M({\tt Age}))= \{\text{Young}, \text{Old}\}$.

\begin{example}
	\label{ex:masked_attributes}
	Consider the {\tt Age} values in Figure~\ref{tab:excerpt_dataset} and a masking function $M$ that returns `Young' for values $10-45$ and `Old' for values $\ge 45$. Then, for masked attribute $M({\tt Age})$ will have values $\langle$ Young, Young, Young, Old, Old, Old $\rangle$.
\end{example}

\begin{table}[t]
    \centering
    \small
    \setlength{\abovecaptionskip}{0mm}
\setlength{\belowcaptionskip}{0mm}
% \captionsetup[subfigure]{aboveskip=0pt,belowskip=0pt}
    \caption{Table of frequently used notations.}
    \begin{tabular}{r|l}
    \toprule
    {\bf Symbol} & {\bf Description} \\
    \midrule
     $M$, $\mM$  & Masking function, Masking configuration \\
     $A$, $M(A)$, $\bY$ & Attribute, Masked attribute, Class label \\
     $D$, $D_\mM$  & Dataset, Masked dataset over configuration $\mM$\\
     $f(a^i)$, $f(a^i,y^j)$  & Frequency of attribute value $a^i$, attribute-label $a^i,y^j$\\
     $\bF(A,\bY)$, $\rho(A;\bY)$ & Joint distribution and Predictive utility measure \\
     \bottomrule
    \end{tabular}
    \label{tab:notations}
\end{table}

\paragraph{Masking Configuration \& Masked Dataset}
A \emph{masking configuration}  is a tuple $\mM = (M_1, M_2, ..., M_m)$, which applies the masking function $M_j$ on attribute $A_j$, where $M_j \in \cC_{A_j}$. Applying $\mM$ to $D$ produces a \emph{masked dataset} $D_\mM = \sset{(x_i', y_i)}_{i=1}^N$, where $x_i' = (M_1(x_{i1}), M_2(x_{i2}), ..., M_m(x_{im}))$. 

\begin{example}
	\label{ex:maskied_datasets}
	Consider two masking configurations that can be used for masking the example dataset of Figure~\ref{tab:excerpt_dataset} (for brevity, we omit other function parameters).
\begin{itemize}
	\item[] $\mM_1$ = (\texttt{{\bf Bu}cketize(Age), {\bf Bl}ur(Weight), {\bf S}uppress(Zipcode)})
	\item[] $\mM_2$ = (\texttt{{\bf G}eneralize(Age),{\bf Bu}cketize(Weight), Zipcode})
\end{itemize}
$D_{\mM_1}$ (below left) and $D_{\mM_2}$ (below right) show the excerpt of masked datasets after applying $\mM_1$ and $\mM_2$to $D$, respectively.
% Left table
\begin{figure}[h]
  \begin{minipage}[b]{0.46\columnwidth}
  \small
  % \vspace{0pt}
    % \centering
    \setlength\tabcolsep{2pt}
    \begin{tabular}{cccr}
      \toprule
      \textbf{B(Age)} & \textbf{Bl(Weight)} & \textbf{S(Zip)} & \textbf{Health} \\
      \midrule
      1-10  & 3* & * & \textbf{G} \\
      1-10  & 3* & * & \textbf{G} \\
      31-40 & 6* & * & \textbf{M} \\
      61-70 & 7* & * & \textbf{P} \\
      71-80 & 8* & * & \textbf{VP} \\
      71-80 & 7* & * & \textbf{VG}\\
      \bottomrule
    \end{tabular}
    % \subcaption{Excerpt of a dataset.}
    % \label{tab:excerpt_dataset}
  \end{minipage}%
  % \hfil
  \begin{minipage}[b]{0.46\columnwidth}
  \small
    % \vspace{0pt}
    \centering
    \setlength\tabcolsep{2pt}
    \begin{tabular}{cccr}
      \toprule
      \textbf{G(Age)} & \textbf{Bu(Weight)} & \textbf{Zip} & \textbf{Health} \\
      \midrule
      Young & 30-35 & 21162 & \textbf{G} \\
      Young & 30-35 & 21168 & \textbf{G} \\
      Young & 60-65 & 22170 & \textbf{M} \\
      Old & 70-75 & 23175   & \textbf{P} \\
      Old & 80-85 & 23173   & \textbf{VP} \\
      Old & 75-80 & 25165   & \textbf{VG}\\
      \bottomrule
    \end{tabular}
    % \subcaption{Excerpt of a dataset.}
    % \label{tab:excerpt_dataset}
  \end{minipage}%
\end{figure}
\end{example}

\paragraph{Predictive Utility Measures of a Dataset}  
The \emph{predictive utility} of a dataset refers to the extent to which it supports accurate predictions in a machine learning (ML) task. In simpler terms, it quantifies how effectively the features (inputs) in the dataset can explain or predict the target variable (label). Various model-specific metrics—such as precision, recall, accuracy, and AUC~\cite{han2012data}—can be used to assess predictive utility. In this paper, however, we focus on model-agnostic, correlation-based measures~\cite{han2012data}, such as mutual information, Chi-Square, and related statistics (see Section~\ref{sec:predutil}).

\paragraph{High-Level Problem}  
In data-sharing ecosystems, the original data is anonymized and never directly accessible. We address the problem of selecting an optimal masking configuration in such settings. In some scenarios, besides the masked predictors, the original marginal distributions of the predictors may or may not be available.\footnote{In practice, frequency distributions are often accessible through histograms or summary statistics.} Given a set of possible masking configurations and a dataset that can only be accessed in its masked form under one of these configurations, the goal is to select the configuration that maximizes the dataset’s predictive utility. Continuing with our running example, the objective is to determine which of two masking configurations—$\mM_1$ or $\mM_2$—yields the highest predictive utility according to the selected measure (e.g., accuracy).

%!TEX root=./main.tex

\section{Correlation-Based Data Masking Advisor} % (fold)
\label{sec:framework_overview}

We propose \framework, a middleware framework designed for data-sharing ecosystems that enables efficient determination of optimal masking configurations. \framework acts as an ``advisor'' to data providers by recommending configurations that maximize predictive utility, independent of any specific downstream task. In the following, we give an overview of \framework and formalize the problem of optimal masking configuration selection that we consider in this paper.

\subsection{\framework Overview}
\label{sub:framework_overview}

\begin{figure}[t]
\captionsetup[subfigure]{aboveskip=0pt,belowskip=0pt}
	\centering
	\includegraphics[page=4]{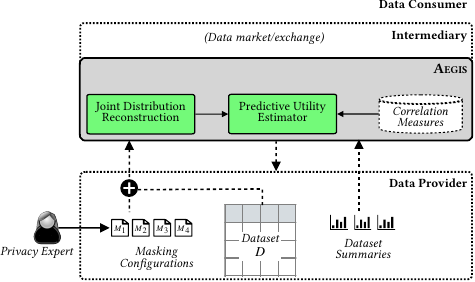}
	\caption{\framework Overview}
    \label{fig:framework_overview}
\end{figure}

Figure~\ref{fig:framework_overview} presents a schematic overview of our proposed framework, \framework. We assume a collaborative setting where data providers work with privacy experts to determine appropriate masking configurations. Each configuration, when applied, produces a masked dataset that meets the minimum privacy guarantees and allows the dataset to be shared within a data-sharing ecosystem. In this context, \framework{} facilitates the identification of an optimal masking configuration \emph{independently} of the downstream task or model, making it particularly useful in scenarios where access to the raw dataset is restricted (e.g., due to privacy regulations). Our approach is grounded in the key principle that the predictive utility of a dataset can be assessed in a model-agnostic manner---i.e., without reliance on specific models---using techniques such as correlation-based measures. To achieve this, \framework is comprises two main components:

\paragraph{Joint Distribution Reconstruction}
As a first step, \framework{} computes the joint distribution of (masked) attribute-label pairs for each masking configuration. When available, it further leverages marginal distributions (i.e., 1D histograms of the dataset's attributes) to reconstruct the joint distribution of \emph{unmasked} attribute-label pairs.

\paragraph{Predictive Utility Estimator}
Given the estimated joint distribution for each (attribute-label, masking configuration) pair\footnote{In practice, the exact masking functions need not be known---an identifier is sufficient.}, the predictive utility estimator computes a \emph{predictive utility deviation} score (formally defined below) for each configuration. The estimator supports well-established, model-agnostic utility measures (see below) to evaluate utility deviation and recommends an optimal masking configuration accordingly.

In what follows, we formally define model-agnostic predictive utility measures, introduce the notion of predictive utility deviation, and describe the problem of selecting an optimal masking configuration that minimizes this deviation.

\subsection{Model Agnostic Predictive Utility}\label{sec:predutil}

Understanding the predictive utility of a dataset while being agnostic about the downstream model or task is crucial in data-sharing settings. In this section, we first explore model-agnostic approaches for estimating predictive utility, focusing on correlation-based measures for discrete and discretized continuous data. We then introduce a novel, model-agnostic metric called Predictive Utility Deviation to quantify utility loss under different masking configurations.

\subsubsection{Predictive Utility Measures}  
The predictive utility of an attribute/feature refers to the amount of information it contributes to predicting the class label $\bY$. This utility is assessed without training a specific machine-learning model in a model-agnostic setting. Instead, statistical or information-theoretic measures such as correlation or association between predictors and the class label are used to estimate how informative the features are. Our framework supports a wide range of measures designed for discrete or categorical data as long as the predictive utility can be captured through the joint distribution of the predictor and the class label. For continuous attributes, we assume that the data has been effectively discretized. Some of the correlation and association measures that we closely study are:

\paragraph{Mutual Information}
The mutual information between predictor $A_j$ and class label \(Y\) is defined as:
\[
I(A; \bY) = \sum_{a^i \in Dom(A)} \sum_{y^j \in Dom(\bY)} P(a^iy^j) \log \left( \frac{P(a^iy^j)}{P(a^{i})P(y^j)} \right)
\]
Where: \(P(a^iy^j)\) is the joint distribution of \(A\) and \(\bY\); and \(P(a^i)\) and \(P(y^j)\) are the marginal probability distributions of \(A\) and \(\bY\), respectively.

\eat{
\begin{itemize}
    \item \(P(a^iy^j)\) is the joint distribution of \(A\) and \(\bY\).
    \item \(P(a^i)\) and \(P(y^j)\) are the marginal probability distributions of \(A\) and \(\bY\), respectively.
\end{itemize}
}
\noindent{\bf $g_3$ Error.}
Given the attribute $A$ and class label $\bY$, $g_3$, the error counts the number of records that must be deleted to satisfy the full functional dependency between $A \rightarrow \bY$.
    \begin{align*}
        g_3(A; \bY)&= \frac{1}{N} (N - max\{ |s|\ |\ s \in \mathcal{D}, \ s \models A \rightarrow \bY\})
    \end{align*}
    In other words, $g_3$ error captures how far a dataset is from satisfying a functional dependency $A\to \bY$ as a functional dependency.

\paragraph{Chi-Square Statistic}
The Chi-Square statistic between attribute $A$ and class label $\bY$ is given by:
\[
\chi^{2} (A; \bY) = \sum_{\forall a^i \in Dom(A), y^j \in Dom(\bY) } \frac{(O_{a^i,y^j} - E_{a^i,y^j})^2}{E_{a^i,y^j}}
\]

Where: \( \chi^2 \) = Chi-Square statistic; \( O_{a^i,y^j} \) = Observed frequency for the \( i,j \)-th domain values of $A$ and $\bY$, respectively; and \( E_{a^i,y^j} \) = Expected frequency for the \( i,j \)-th domain values of $A$ and $\bY$, respectively.
 
\eat{ 
\begin{itemize}
    \item \( \chi^2 \) = Chi-Square statistic.
    \item \( O_{a^i,y^j} \) = Observed frequency for the \( i,j \)-th domain values of $A$ and $\bY$, respectively.
    \item \( E_{a^i,y^j} \) = Expected frequency for the \( i,j \)-th domain values of $A$ and $\bY$, respectively.
\end{itemize}
}

\begin{example}
Continuing the running example for the dataset in Figure~\ref{fig:example_dataset}(a), we have $I(\mathrm{Age};\mathrm{Health}) \approx 0.640$; $g_{3}(\mathrm{Age};\mathrm{Health}) = 0.53$; and $\chi^2(\mathrm{Age};\mathrm{Health}) \approx 85.96$.    
\end{example}

\subsubsection{Predictive Utility Deviation}
For a given dataset $D$, a feature/predictor $A$, and a class label $\bY$ let $\rho$ be the predictive utility measure (e.g., Mutual Information, Chi-Square, etc.), where $\rho(D, A,\bY)$ returns a scalar value that reflects the predictive utility of $A$ for $\bY$.
 If the $m$ predictors in $D$ are masked using a masking configuration $\mM = (M_1, M_2, ..., M_m)$, we define the \emph{predictive utility deviation} (PUD) as the sum of the absolute differences between the predictive utility computed on the original dataset $D$ and that of the masked dataset $D_\mM$:
\begin{equation}\label{eq:pud}
	\Delta_{PU}(D,D_\mM, \rho) = \frac{1}{m}\sum_{i=1}^m \sz{\rho(A_i;\bY) - \rho(M(A_i);\bY)}
\end{equation}

\subsection{Problem Definition} % (fold)
\label{sub:problem_definition}

\begin{problem*}
{\bf (Identify Optimal Masking Configuration to Minimize Predictive Utility Deviation.)} Given a dataset $D$, a predictive utility measure $\rho$, and a set $\cM_{set} = \sset{\mM_1,\mM_2,\dots,\mM_K}$	of $K$ predefined masking configurations, find the masking configuration $\mM^* \in \cM_{set}$ that minimizes predictive utility deviation, i.e.,
\[
\mM^* = \argmin_{\mM \in \cM_{set}} \Delta_{PU}(D, D_\mM, \rho)
\]
\end{problem*}

The primary challenge is that to compute any predictive utility measure $\rho$, the joint frequency distribution between each $A$ and $\bY$ must be known. When the raw data $D$ itself is not accessible, the joint distributions between the predictors and the class label are unknown, so $\Delta_{PU}$ (Equation~\ref{eq:pud}) cannot be directly calculated from the data. We study two variants of this problem to that end.

\begin{problem*}
{\bf (Variant I: Identify optimal masking configuration in the presence of marginal distributions.)} Given a dataset $D$ where the joint distributions of the attributes and the class label are not accessible, but the marginal distributions $Dist(A)$ of each masked predictor $A$ is available. A set $\cM_{set} = \sset{\mM_1,\mM_2,\dots,\mM_K}$	of $K$ predefined masking configurations, for a given predictive utility measure $\rho$, find the masking configuration $\mM^* \in \cM_{set}$ that minimizes the predictive utility deviation.
\end{problem*}

\begin{problem*}
{\bf (Variant II: Identify optimal masking configuration in the absence of marginal distributions.)} Given a dataset $D$ where neither marginals nor joint distributions are accessible, for a given predictive utility measure $\rho$, and a given set $\cM_{set} = \sset{\mM_1,\mM_2,\dots,\mM_K}$	of $K$ predefined masking configurations, find the masking configuration $\mM^* \in \cM_{set}$ that minimizes the predictive utility deviation.
\end{problem*}

%!TEX root=./main.tex

\section{Algorithms inside \framework{}} % (fold)
\label{sec:reconstructing_join_distributions}

We now turn attention to the algorithms inside \framework{}. We first describe the algorithm for selecting the masking configuration with the minimum predictive utility deviation (Section~\ref{sub:overall_algorithm}). We will then discuss how \framework{} reconstructs joint distributions to estimate the utility of masked datasets (Sections~\ref{sub:reconstructing_joint_distributions} and \ref{sub:predictive_utility_deviation}).

\subsection{Overall Algorithm} % (fold)
\label{sub:overall_algorithm}

Algorithm~\ref{algo:best_masking} outlines the end-to-end process for selecting the optimal masking configuration. For each configuration $\mM_k$, the algorithm iterates over every masked predictor–label pair $\bigl(M(A),\bY\bigr)\in\mM_k$ and first reconstructs the original joint distribution $\bF(A,\bY)$ by invoking Subroutine {\tt Reconstruction} (Algorithm~\ref{alg:recon}, line4). Then, using a chosen correlation or association measure $\rho$ (see Section~\ref{sec:predutil}), it computes the predictive utility deviation between the masked distribution $\bF(M(A),\bY)$ and the reconstructed distribution $\bF(A,\bY)$ (lines~5–7). These individual deviations are averaged across all masked attributes in $\mM_k$ to yield the total predictive utility deviation as defined in Equation~\ref{eq:pud} (line8). Finally, the configuration with the smallest total deviation is selected (line~9). Through this sequence of reconstruction, deviation computation, and aggregation, Algorithm~\ref{algo:best_masking} allows identifying the masking configuration that minimizes loss of predictive information. It is worth noting computing $\bF(M(A),\bY)$ does not require materializing the masked dataset for each configuration.

\begin{algorithm}[t]
% \setlength{\abovecaptionskip}{0mm}
% \setlength{\belowcaptionskip}{0mm}
% \captionsetup[subfigure]{aboveskip=0pt,belowskip=0pt}
\caption{Finding the Best Masking Configuration}
\label{algo:best_masking}
\begin{algorithmic}[1]
\REQUIRE Set of attributes $\mathcal{A}$, target $\bY$, set of masking configurations $\cM_{\text{set}}$, available constraints, predictive utility measure $\rho$
\STATE Initialize $\mM^* \gets \text{None}$
\FOR{each $\mM_k \in \cM_{\text{set}}$}
    \FOR{each attribute $A$}
        \STATE $\bF(A,\bY)$  $\leftarrow$ Subroutine {\tt Reconstruction}(A, constraints)
        \STATE Compute predictive utility $\bF(A,\bY),\rho)$ 
        \STATE Compute predictive utility $\bF(M(A),\bY),\rho)$ 
        \STATE Compute predictive utility deviation $(\bF(A,\bY), \bF(M(A),\bY),\rho)$ 
        \ENDFOR
        \STATE Compute total predictive utility deviation of $\mM_k$.
        \ENDFOR
\STATE $\mM^* \gets$ configuration with the smallest predictive utility deviation.
\RETURN $\mM^*$
\end{algorithmic}
\end{algorithm}

\begin{lemma}\label{lemma:Alg1_time}
Running time of Algorithm~\ref{algo:best_masking} is $\mathcal{O}(|\mM_k| \times |\cM_{\text{set}}| \times \bT)$, where $\bT$ is the time it takes to run Subroutine {\tt Reconstruction} once.
\end{lemma}

\subsection{Reconstructing Joint Distributions}
\label{sub:reconstructing_joint_distributions}

We now describe reconstructing the joint distribution, i.e., given a masking function $M$, a masked attribute $M(A)$, label $\bY$, and how to reconstruct $\bF(A,\bY)$.

\begin{algorithm}[t]
\caption{Subroutine {\tt Reconstruction}}\label{alg:recon}
\begin{algorithmic}[1]
\REQUIRE Uniform joint distribution $\bF_{uni}(A,\bY)$, constraints on rows of $\bF(A, \bY)$, constraints on columns of $\bF(A, \bY)$, convergence criteria
\ENSURE Adjusted joint distribution $\bF(A, \bY)$
\STATE Initialize $\bF(A, \bY) \leftarrow \bF_{uni}(A,\bY)$ 
\REPEAT
    \FOR{each row $i$}
        \STATE Adjust $\bF(A, \bY)_{i, :}$ such that row constraints are satisfied
    \ENDFOR
    \FOR{each column $j$}
        \STATE Adjust $\bF(A, \bY)_{:, j}$ such that column constraints are satisfied
    \ENDFOR
\UNTIL{Convergence criteria is met}
\RETURN $\bF(A, \bY)$
\end{algorithmic}
\end{algorithm}

The masked attribute $M(A)$ and the masking function $M$ together reveal more information: imagine the following simple example: the attribute {\em Age} is masked, and the masked data contains 20 young and 80 old individuals. The masking function $M$ imposes further constraints: {\em Young  $\rightarrow 10-45$, Old $\rightarrow > 45$}. From this, the reconstructed Age has to satisfy the following:  $20$ individuals of Age between $10-45$, $80$ individuals are above $45$. Additionally, when the marginal of {\em Age}, i.e. $Dist(A)$ is known, that imposes further constraints. The challenge, however, is that a prohibitively large number of possible distributions of {\em Age} satisfies all these constraints, and any of these could be a likely solution. Without any additional information, we assume that these values are equally likely. We formalize this as an optimization problem, i.e., estimate the joint distribution of each predictor and the class label (i.e., {\em Age} and {\em Health}) such that the distribution of the predictor  (i.e., {\em Age}) is as uniform as possible within the given range. Still, the reconstructed joint distribution satisfies all constraints. 

When the marginal distribution of the attribute is known, i.e., $Dist(A)$ is available, further information is known. For example, for predictor {\em Age}, there exists $4$ individuals with age $10$, $12$ individuals $17$, $4$ with $43$, and so on. 

\noindent {\bf Constraints.} Formally speaking, to be able to estimate the joint distribution of attribute $A$ and the class label $\bY$, i.e., $\bF(A,\bY)$, the reconstruction process must satisfy several constrains ``implied'' by the input: the marginal distribution of $A$ (denoted $Dist(A)$; the masking function $M$ together with the resulting masked attribute $M(A)$; the joint distribution of masked $A$ and $\bY$, i.e., $\bF(M(A),\bY)$; and the total number of records $N$.

\begin{itemize}
    \item Type 1: $Dist(A)$ is known, it imposes the following set of constraints: the frequency of each unique domain value $a^i$ of actual $A$, i.e., \(a^i _{\in Dom(A)} f(a^i)\). 
    \item The masking function $M$ reveals, if the masked value of the predictor is $a^{i'}$ what is the likely set of $a^{i}$ values, i.e., a mapping between $a^{i'} \rightarrow \{a^i\}$, $\forall a^{i'} \in Dom(M(A))$. 
    \item Type 2: The masking function $M$ and masked attribute $M(A)$ thus imposes an additional set of constraints: the frequency of each unique domain value of the masked attribute $A$: i.e.,
    $a^{i'}_{\in Dom(M(A))} f(a^{i'})$.
    \item Type 3: The joint distribution  of masked $M(A)$ and $\bY$, i.e., $\bF(M(A),\bY)$. This joint distribution $\bF(M(A),\bY)$ imposes another set of constraints, the frequency of each unique \\ ${i',j}$,  $a^{i'}_{\in Dom(M(A))},y^j_{\in Dom(\bY)}f(a^{i'}y^j)$. 
    \item Finally, the total number of records $N$ is known.
    \end{itemize}
\noindent The task is to estimate the joint distribution $\bF(A,\bY)$ of a predictor $A$ and class label $\bY$ from these above known constraints, such that all constraints are satisfied and $\bF(A,\bY)$ is as uniform as possible. 

To be able to formalize the optimization problem, let $\bF_{uni}(A,\bY)$ be the joint distribution where $\bF_{uni}(A,\bY)$ is fully uniform. This distribution has $|Dom(A)|$ number of $A$ values and $|Dom(\bY)|$ number of $\bY$ values. Since this distribution is fully uniform, $N$ is equally divided between $|Dom(A)| \times |Dom(\bY)|$ number of possibilities. 

\begin{example}
Let $N$ be $100$, $|Dom(A)|=4$,$|Dom(\bY)|=5$, $|Dom(A)| \times |Dom(\bY)|=20$, then $\bF_{uni}(A,\bY)$ should initially ``look'' like: \\
\begin{center}\small
\begin{tabular}{|c|c|c|c|c|c|}
\hline
\textbf{} & \textbf{Very Poor} & \textbf{Poor} & \textbf{Moderate} & \textbf{Good} & \textbf{Very Good} \\
\hline
\textbf{Age10} & 5 & 5 & 5 & 5 & 5 \\
\textbf{Age12} & 5 & 5 & 5 & 5 & 5 \\
\textbf{Age17} & 5 & 5 & 5 & 5 & 5 \\
\textbf{Age43} & 5 & 5 & 5 & 5 & 5 \\
\hline
\end{tabular}
\end{center}
\end{example}

When no constraints are imposed, the ``maximally uninformative’’ choice for the joint distribution of $(A,\bY)$ is the uniform law
\[
\bF_{uni}(A,\bY)\;=\;\frac N{|\Dom(A)|\;\,|\Dom(\bY)|}
\quad\forall\,a\in\Dom(A),\;y\in\Dom(\bY)\,.
\]  
However, this $\bF_{uni}$ will almost certainly violate the marginals (of $A$ and $M(A)$), the joint $(M(A),\bY)$ frequencies, and the total count $N$.  To enforce all of those constraints, we must perturb $\bF_{uni}$ into some new distribution $\bF$ that exactly matches every input statistic.  Yet among the infinitely many distributions satisfying those linear constraints, we still wish to remain as ``close'' as possible to the original uniform prior, i.e. to introduce no additional bias beyond what the constraints demand. The objective function could be formally written as follows:

\begin{equation*}\label{eq:obj}
\begin{aligned}
&{\bf Minimize} \quad \text{DIST}\left( \bF_{\text{uni}}(A, \bY),\ \bF(A, \bY) \right) \text{ \bf subject to }\\
& \quad \text{satisfy constraints } f(a^i), \quad a^i \in \text{Dom}(A) \\
&  \quad  \text{satisfy constraints } f(a^{i'})\quad \forall a^{i'} \in \text{Dom}(M(A)) \\
&  \quad \text{satisfy constraints }  f\left(a^{i'}y^{j}\right), \forall a^{i'} \in \text{Dom}(M(A)),\ \forall y^j \in \text{Dom}(\bY).
\end{aligned}
\end{equation*}

\noindent The concept of distance \text{DIST} between the two joint distributions can be formalized in multiple ways, depending on the choice of the norm used to measure the difference. It could be $\ell_2$ norm $ \sqrt{\Sigma_{\forall i, j} [\bF_{\text{uni}}(A, \bY)_{i,j}- \bF(A, \bY)_{i,j}]^2}$, which is the Euclidean distance between the matrices, treating them as flattened vectors. This is heavily used in practice due to its nice mathematical properties (differentiable, smooth).
On the other hand, $\ell_1$ norm,  $\Sigma_{\forall i, j} |\bF_{\text{uni}}(A, \bY)_{i,j}- \bF(A, \bY)_{i,j}|$, is the sum of absolute differences between corresponding entries of the two distribution. This measure is more robust to outliers compared to $\ell_2$. Finally, $\ell_{\infty}$ norm $Max_{\forall i, j} |\bF_{\text{uni}}(A, \bY)_{i,j}- \bF(A, \bY)_{i,j}|$ captures the worst case deviation between the two joint distributions.\looseness=-1

If the problem is formalized in $\ell_2$ this gives rise to an Integer Quadratic Programming problem, which is non-convex and NP-hard in general~\cite{wolsey1999integer}. $\ell_2$  error, defined as the sum of squared errors, is the most popular error function because of its key mathematical and statistical properties~\cite{bjorck1990least}. It leads to simpler and faster parameter estimation due to the availability of closed-form solutions. This makes it computationally efficient, particularly for large datasets. Statistically, in the presence of Gaussian noise, $\ell_2$ corresponds to the maximum likelihood estimator. Another key advantage of $\ell_2$  error is that its error function is differentiable, allowing for the application of calculus-based optimization techniques such as gradient descent. We have updated the revision accordingly.

Instead of solving using integer quadratic programming, we adapt iterative proportional fitting (IPF)\cite{idel2016review, kruithof1937calculation} to solve this problem efficiently. See subroutine {\tt Reconstruction} in Algorithm~\ref{alg:recon}. It starts with a uniform joint distribution $\bF_{uni}(A,\bY)$. The constraints derived from known masking functions masked 2-D distributions and unmasked 1-D distribution are used to impose a set of row (see Type~1 constraint above) and column (cf.~Type~2 and 3 above) constraints on $\bF(A, \bY)$. The algorithm takes turns and readjusts rows to satisfy all row-related constraints. Then, it returns, takes turns, and readjusts columns to satisfy all constraints. Adjusting row constraints will likely perturb the column constraints (and vice versa). This process is repeated iteratively until $\bF(A, \bY)$  converges —i.e., the adjusted values closely align with the target marginals within a specified tolerance. 

It is known that all the observed values in the 2-D data table are strictly positive (greater than zero)~\cite{ruschendorf1995convergence}, then the existence and uniqueness of the maximum likelihood estimates are ensured, and thus convergence of IPF is guaranteed. For our problem, the underlying joint frequency distribution table satisfies this condition, hence IPF converges in our case.

In general, however, IPF  converges to fit marginal totals within a selected level of closeness. The algorithm typically uses criteria such as the maximum number of iterations or a convergence criterion (e.g., largest proportional change in an expected cell count) to determine when to stop iterating. Lowering the tolerance for convergence will require more iterations.

A challenge still remains is that the Subroutine {\tt Reconstruction} can give fractional numbers in the joint distribution, whereas the reconstructed joint distribution must only contain integral values. To address this issue and enforce integrality, we apply a randomized rounding procedure as a post-processing step at the end of the \texttt{Reconstruction} subroutine. In this step, each fractional value in the joint distribution is normalized and interpreted as the probability of rounding up to the next integer (or down to the integer before), and independent random trials are used to generate integer entries. This ensures that the final distribution consists entirely of integers while preserving, in expectation, the original fractional values. Exploring the approximation factor of this randomized rounding~\cite{raghavan1987randomized} is left to future work.

\noindent {\bf Running Time Analysis.} Running time of Subroutine {\tt Reconstruction} is $\mathcal{O}(\#rows \times \#columns)$ of $\bF(A, \bY)$ per iterations. Let $\bI$ denote that time. If the process takes $t$ iterations for convergence, the total running time $\bT$ is $\mathcal{O}(t \times \bI)$.

\noindent {\bf Case I: 1D Histogram Known}
When ID histograms are known, Subroutine {\tt Reconstruction} is run, and the input constraints are of Type 1, Type 2, and Type 3. 

\noindent {\bf Case II: 1D Histogram Unknown}
When the marginal of the predictors are unknown, Subroutine {\tt Reconstruction} is run, but the input constraints are of only Type 2 and Type 3. Clearly, with fewer constraints, the subroutine converges faster.

\subsection{Predictive Utility Estimation}
\label{sub:predictive_utility_deviation}

Our algorithm to estimate the predictive utility deviation of a masking configuration $\mM_k$ takes a predictive utility measure $\rho$ as an input,  as well as two joint distributions: $\bF(M(A),\bY)$ is the masked joint distribution, and $\bF(A,\bY)$ is the reconstructed joint distribution from Subroutine {\tt Reconstruction} for each attribute $A$. It then computes the predictive utility deviation of $\mM_k$. The masking configuration with the smallest deviation is then determined as the optimal masking configuration.
  
\begin{example}
Consider the masked joint distribution $\bF(M(A),\bY)$ of {\em Age, Health} as shown below. \\
\begin{center}\small
\begin{tabular}{|c|ccccc|c|}
\hline
\textbf{$M$(Age) $\downarrow$ / Health $\rightarrow$} 
  & \textbf{VP} & \textbf{P} & \textbf{M} & \textbf{G} & \textbf{VG} & \textbf{Row Total} \\
\hline
\textbf{Young (10–45)} &  0 &  0 &  1 &  7 & 12 & 20 \\
\textbf{Old  (>45)}    & 15 & 25 & 29 &  9 &  2 & 80 \\
\hline
\textbf{Column Total}  & 15 & 25 & 30 & 16 & 14 & 100 \\
\hline
\end{tabular}
\end{center}

\noindent
Then for mutual information (MI) as the predictive utility measure, we have $p(\mathrm{Young})=0.2$, $p(\mathrm{Old})=0.8$, $p(\mathrm{Very Poor})=0.15$, $p(\mathrm{Poor})=0.25$, $p(\mathrm{Moderate})=0.30$, $p(\mathrm{Good})=0.16$, $p(\mathrm{Very Good})=0.14$. Continuing the process one can show, $I(\mathrm{M(Age);} \mathrm{Health}) \approx 0.42$

\end{example}

Similarly, its estimated mutual information could be calculated from the reconstructed {\em Age, Health}. The mutual information deviation is the absolute difference between these two.

%!TEX root=./main.tex

\section{Evaluation} % (fold)
\label{sec:evaluation}

We experimentally evaluate \framework on real-world datasets to investigate the (a)~overall efficiency and effectiveness of the framework in determining optimal masking configurations; (b)~quality of reconstructed joint distributions in the presence and absence of data summaries; (c)~how the number of attributes, masking configurations, and dataset size impact the estimation; and (d)~scalability of the framework.

Our evaluation demonstrates that \framework can determine an optimal masking configuration order of magnitude faster while achieving comparable predictive performance for various downstream ML tasks. Our IPF-based reconstruction of joint distributions using both available and unavailable data summaries yields superior quality compared to the sampling-based approach. The parameters (number of attributes, configurations, and rows) do not impact \framework's effectiveness, and our approach scales nearly linearly with an increase in the number of attributes, configurations, and dataset size. Overall, we found \framework{} an efficient and effective middleware framework in data-sharing ecosystems for selecting masking configurations.\looseness=-1

\subsection{Experimental Setup} % (fold)
\label{sub:experimental_setup}

\subsubsection{Implementation and Setup}
We implemented all algorithms in \framework using Python~3.11.7. All experiments were performed on a server, which is equipped with Intel i9-11900K\@3.50GHz with 8 CPU cores and 128GB RAM. Additional scalability experiments (Section~\ref{sub:framework_efficiency}) were conducted on server with AMD EPYC 7713 CPU (2.0GHz), 128 cores, 512GB RAM, and a single NVIDIA A100 GPU with 80GB memory.

%!TEX root=./main.tex

\begin{table*}[t]
% \setlength{\abovecaptionskip}{0mm}
% \setlength{\belowcaptionskip}{0mm}
% \captionsetup[subfigure]{aboveskip=0pt,belowskip=0pt}
\scriptsize
\centering
\caption{Overview of benchmark dataset collection and features' statistics. Cat.:Categorical; Num.: Numerical; DS: Domain Size (for Num. Attr.); Min./Max./Avg. \#categories (catgr) across Cat. attributes; Min./Max./Avg. range over Num. attributes.}
\label{tab:datasets_overview}
\begin{tabular}{llp{0.35cm}p{0.5cm}p{0.5cm}p{0.5cm}p{0.5cm}p{0.5cm}p{0.5cm}p{0.5cm}p{0.3cm}ll}
\toprule
\textbf{Name}  & \textbf{\#Rows}                     & \textbf{\#Cat.} & \textbf{\#Num.} & \textbf{Min. DS} & \textbf{Max. DS} & \textbf{Avg. DS} & \textbf{Min. \#catgr.} & \textbf{Avg. \#catgr.} & \textbf{Max. \#catgr.} & \textbf{Min. range} & \textbf{Avg. range} & \textbf{Max. range} \\
\midrule
Air Quality~\cite{airquality} (AQ) & 5,000 & 1  & 9  & 4  & 683   & 141.78  & 4 & 4     & 4    & 3.0  & 183.78        & 769.00 \\
Customer~\cite{hotelbooking} (HCI) & 10,000 & 10 & 15 & 5  & 10000 & 2428.40 & 2 & 657   & 3245 & 4.0  & 1.43$\times10^9$ & 1.06$\times 10^{10}$  \\
Income~\cite{incomedata} (IN)  & 32,561     & 9  & 6  & 16 & 21648 & 3673.67 & 2 & 11.56 & 42   & 15.0 & 262826.83     & 1472420.00\\
\bottomrule
\end{tabular}
\end{table*}

\subsubsection{Datasets}
We used three real-world datasets from Kaggle~\cite{airquality, hotelbooking, incomedata} pertaining to environmental monitoring, e-commerce, and socioeconomic domains. The datasets are summarized in Table~\ref{tab:datasets_overview}.\\[-1.5em]
\begin{itemize}
\item\textbf{Air Quality Dataset.}~\cite{airquality} This data set captures environmental and demographic indicators associated with exposure to pollution. The prediction task is a four-class classification problem based on air quality labels.

\item\textbf{Customer–Hotel Interaction Dataset.}~\cite{hotelbooking} This data set contains features that describe user engagement and hotel characteristics. The target is a binary indicator of booking success.

\item\textbf{Income Dataset.}~\cite{incomedata} This benchmark dataset includes sensitive socioeconomic variables such as age, education level, and employment status. It supports a binary income classification task.\looseness=-1
\end{itemize}

In addition, we also used synthetic datasets to evaluate \framework’s scalability under controlled variations in dataset size, number of attributes, and number of masking configurations (see Section~\ref{sub:framework_efficiency}).

\subsubsection{Data Summaries}

For each attribute in the dataset, we construct one-dimensional histograms from which the marginal distribution of attribute values can be derived. We encode each histogram as a dictionary that maps unique discrete values to their corresponding frequencies. Further, to evaluate the impact of available summaries on joint distribution reconstruction quality, we consider two settings.

\begin{itemize}
    \item \textbf{\textsc{Aegis}+1D}: In this setting, the marginal distributions of the individual attributes are available and used during reconstruction (recall case~1; Section~\ref{sub:reconstructing_joint_distributions}). These histograms provide statistical guidance that improves the accuracy of estimating joint distributions.\looseness=-1

    \item \textbf{\textsc{Aegis}-1D}: In this setting, the reconstruction is performed without access to attribute-level statistics, and we assume a uniform prior over each attribute's domain (recall Case~2; Section~\ref{sub:reconstructing_joint_distributions}).
\end{itemize}

In the following, unless otherwise stated, we assume access to data summaries as the default setting.

\subsubsection{Masking Configurations}

To evaluate \framework under diverse privacy scenarios, we develop a configurable masking configuration generator that applies a range of privacy-preserving transformations to the dataset. The generator supports both generalization-based masking functions (e.g., bucketization, generalization, and blurring) and attribute suppression, with parameterizable controls to vary the extent of information loss. The generator constructs multiple masking configurations by carefully assigning masking functions to attributes while ensuring the target variable remains unmodified. We used the set of 50 masking configurations across both reconstruction settings (with and without access to data summaries) to ensure a consistent basis for comparison for each dataset. In addition, we used  privacy-preserving masking configuration based on $k$-anonymization using ~\cite{sainzpardo2024anjana} (see Section~\ref{ssub:effectiveness_on_k_anonymization_based_configurations})

\subsubsection{Correlation Measures and Downstream Tasks.}
We evaluated \framework{} using correlation measures from Section~\ref{sec:predutil} including $g_3$, mutual information (MI), and Chi-square test ($\chi^2$). For downstream tasks, we consider several ML models, including Logistic Regression (LR), Support Vector Machine (SVM), Random Forest (RF), Stochastic Gradient Descent (SGD), and Naive Bayes (NB).

\subsubsection{Competitors}

\paragraph{Baseline} We implemented the traditional approach (recall Section~\ref{sec:introduction}) as a baseline that exhaustively applies each masking configuration to first obtain a masked dataset on which a downstream model is trained. This allows us to compute the gold standard accuracy, which we use as our predictive utility metric. 

\paragraph{\textsc{Mascara}} We also compared to \textsc{Mascara}~\cite{mascara}, a recently proposed middleware system for disclosure-compliant query answering. While \textsc{Mascara} is designed as a SQL rewriting system in the presence of a masking function, we adapt its utility estimator to our setting to determine optimal masking configurations.

\paragraph{\textsc{Sampling}}
We also implemented a sampling-based reconstruction in \textsc{Aegis} that generates synthetic records by sampling from original data in accordance with masked attribute ranges. For each attribute in the masked data, we identify the possible set of values and then randomly sample from it with replacement. This procedure produces a reconstructed dataset where the distribution of attribute values matches the masked view, and the associated labels are directly inherited from the sampled original records.

\subsection{Overall Efficiency and Effectiveness} % (fold)
\label{sub:effectiveness}

\subsubsection{End-to-end Efficiency}
\begin{figure*}[t]
% \setlength{\abovecaptionskip}{0mm}
% \setlength{\belowcaptionskip}{-3mm}
% \captionsetup[subfigure]{aboveskip=0pt,belowskip=0pt}
  \centering
  \begin{subfigure}{0.48\linewidth}
    \centering
    \includegraphics[width=\linewidth]{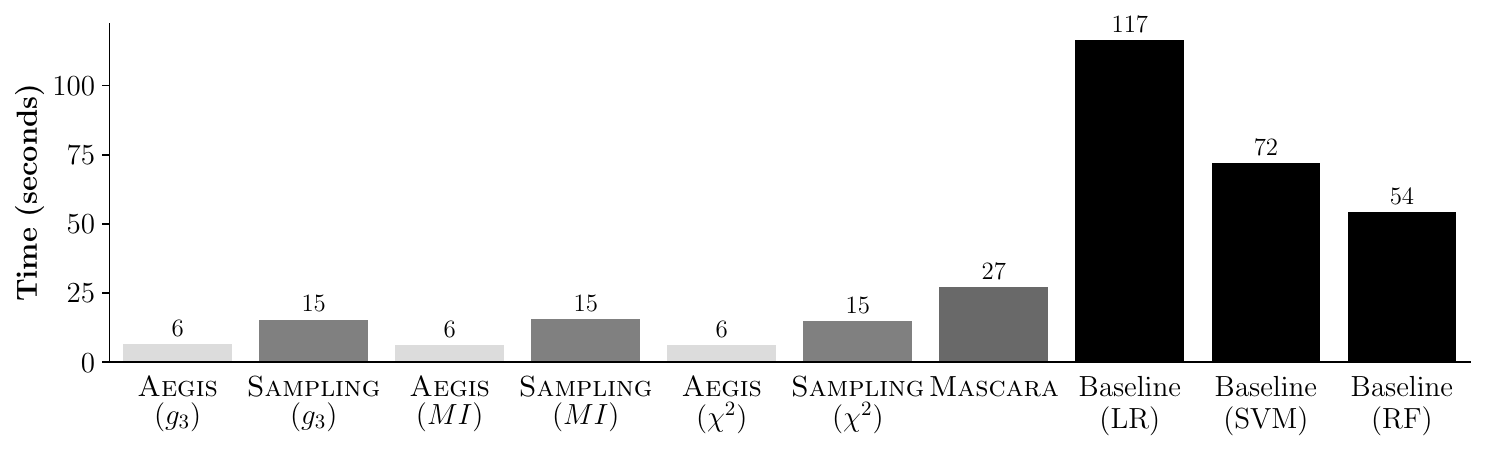}
    \caption{AQ Dataset}
    \label{fig:runtime_dataset_1}
  \end{subfigure}\hfill
  \begin{subfigure}{0.48\linewidth}
    \centering
    \includegraphics[width=1\linewidth]{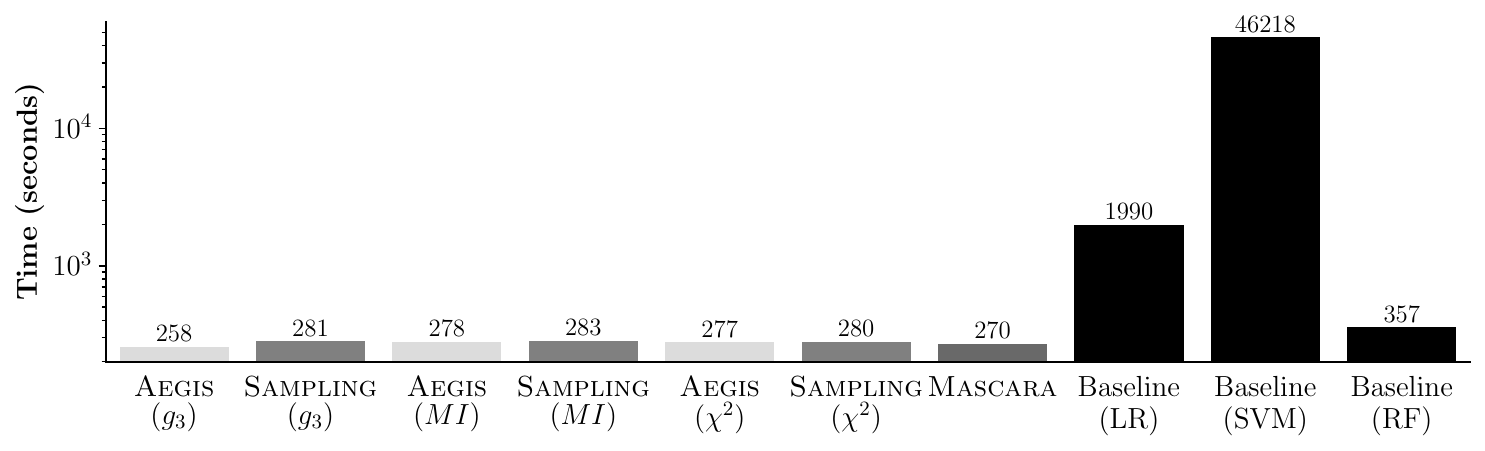}
    \caption{HCI Dataset}
    \label{fig:runtime_dataset_2}
  \end{subfigure}
  
  % \vspace{0.5em} % spacing between subfigures

  \begin{subfigure}{0.48\linewidth}
    \centering
    \includegraphics[width=1\linewidth]{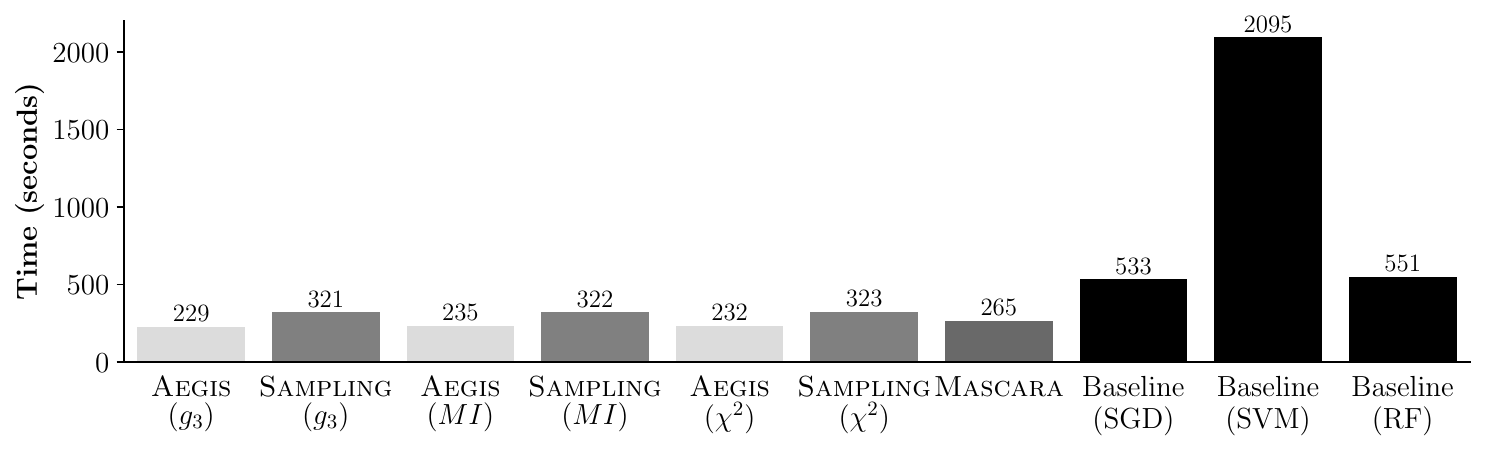}
    \caption{IN Dataset}
    \label{fig:runtime_dataset_3}
  \end{subfigure}\hfill
  \begin{subfigure}{0.48\linewidth}
    \centering
    \includegraphics[width=1\linewidth]{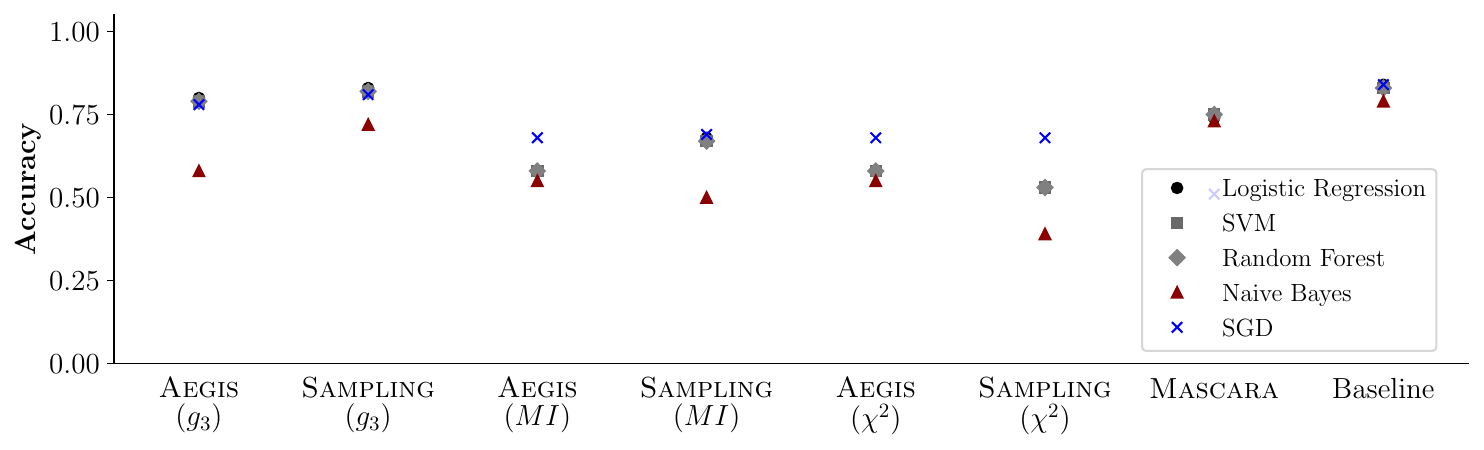}
    \caption{AQ Dataset}
    \label{fig:accuracy_d1_all}
  \end{subfigure}

  % \vspace{0.5em} % spacing between subfigures

  \begin{subfigure}{0.48\linewidth}
    \centering
    \includegraphics[width=1\linewidth]{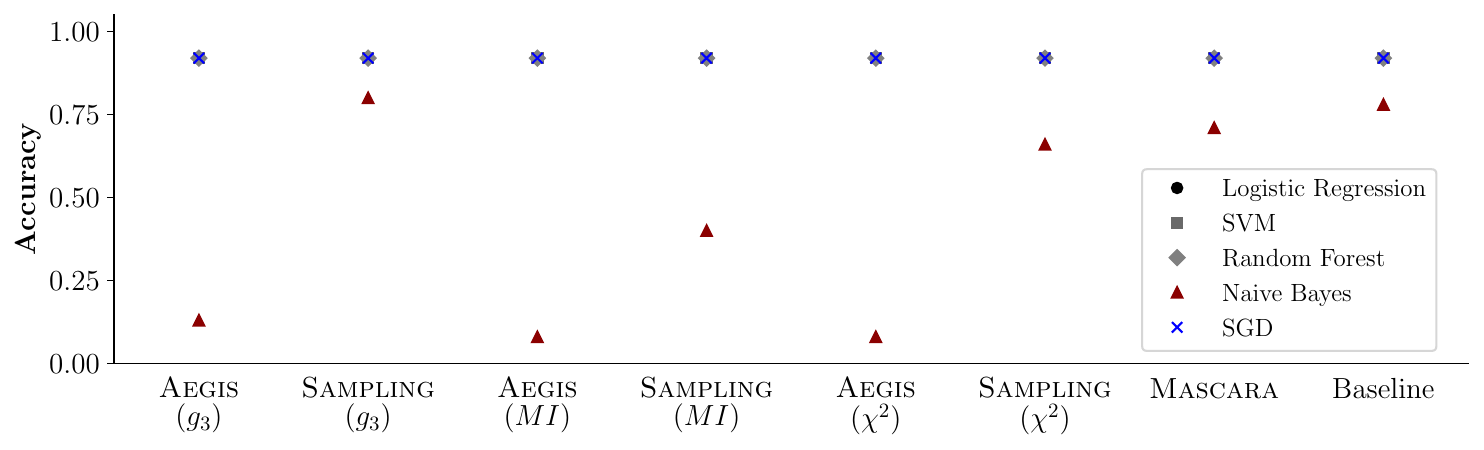}
    \caption{HCI Dataset}
    \label{fig:accuracy_d2_all}
  \end{subfigure}\hfill
  \begin{subfigure}{0.48\linewidth}
    \centering
    \includegraphics[width=1\linewidth]{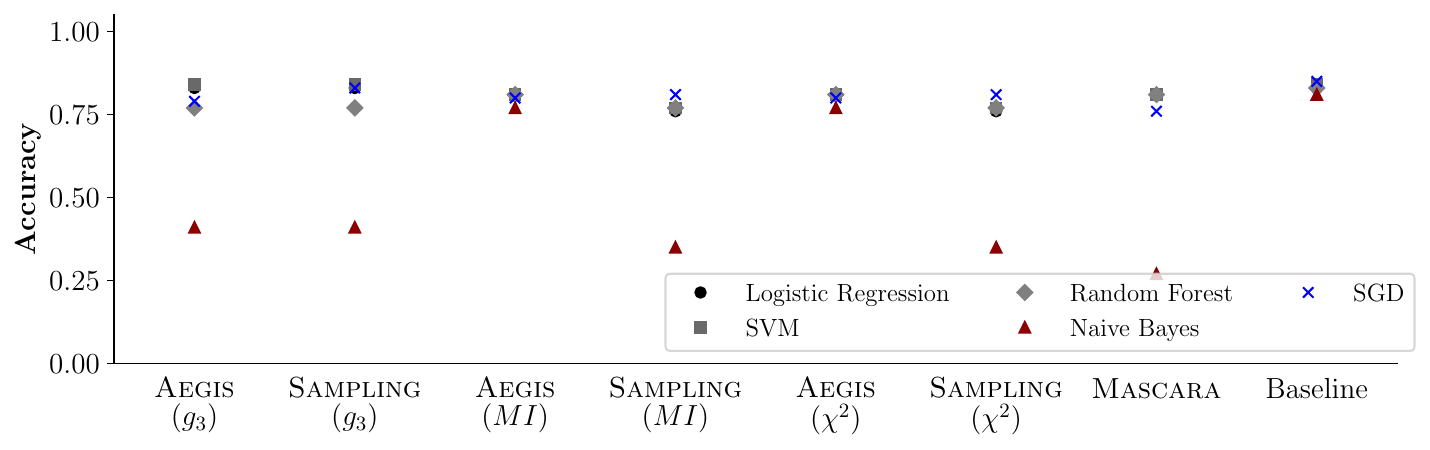}
    \caption{IN Dataset}
    \label{fig:accuracy_d3_all}
  \end{subfigure}

  \caption{Overall efficiency (a)-(c) and effectiveness (d)-(f) of \framework.}
  \label{fig:end_to_end}
\end{figure*}

Figures~\ref{fig:runtime_dataset_1}-\ref{fig:runtime_dataset_3} compares the end‑to‑end runtime of \framework against both \textsc{Sampling}, \textsc{Mascara}, and the baseline approach across the three datasets. We consider logistic regression, SVM, and random forest as the downstream tasks. 

On the smaller AQ dataset (Figure~\ref{fig:runtime_dataset_1}), \framework completes the reconstruction and predictive‑utility evaluation in just over $6s$ for each of the three measures (\(g_3\), MI, and $\chi^2$), yielding a 2.5$\times$ speedup over the corresponding \textsc{Sampling} ($\approx15$s) and a $4\times$ improvement over \textsc{Mascara} ($27s$). In contrast, the baseline approach incurs substantially higher cost---$116s$ for logistic regression, $72s$ for SVM, and $54s$ for random forest---demonstrating that \framework can evaluate masking configurations nearly an order of magnitude faster.   

On the larger HCI and IN datasets (Figures~\ref{fig:runtime_dataset_2} and \ref{fig:runtime_dataset_3}), \framework remains competitive with both \textsc{Sampling} and \textsc{Mascara}. At the same time, it outperforms the baseline by multiple orders of magnitude. For example, on HCI, the \framework is over 7$\times$ faster for logistic regression and more than 160$\times$ for SVM. For IN Dataset, it is roughly 2$\times$ faster for both SGD ($533s$) and random forest ($551s$) and nearly $9\times$ faster than SVM ($2,095s$). 

\emph{Overall, \framework provides an efficient framework to determine optimal masking configuration.}

\subsubsection{End-to-end Effectiveness}
\label{ssub:end_to_end_effectiveness}
Figures~\ref{fig:accuracy_d1_all}-\ref{fig:accuracy_d3_all} compares the predictive utility of masked datasets (AQ, HCI, and IN) using five classifiers (logistic regression, random forest, naive Bayes, and SGD). We compare \framework (using $g_3$, MI, and $\chi^2$ measures) with \textsc{Sampling}, \textsc{Mascara}, and the baseline approach that serves as the gold standard. \looseness=-1

As general trends, we observe that classifiers trained on masked data (determined by \framework) achieve accuracies close to the baseline.
In particular, using $g_3$ error as a correlation measure consistently yields the best masking configuration, with an accuracy loss of less than 3--7\% for all classifiers except Naive Bayes (see discussion below). On average, \framework achieves accuracy comparable to the \textsc{Sampling} approach and outperforms \textsc{Mascara}, which adapts KL-divergence as a utility metric. In addition, except for Naive Bayes, all classifiers had an average accuracy of 0.69, 0.92, and 0.8 for AQ, HCI, and IN datasets.

More specifically, on the AQ dataset (Figure~\ref{fig:accuracy_d1_all}), we observe that for Logistic Regression (circle marker), using $g_3$ error as a correlation measure yielded the best accuracy for both \framework (0.81) and \textsc{Sampling} (0.82), compared to the baseline (0.84). For other correlation measures, both \framework and \textsc{Sampling} performed up to 30\% worse than the baseline, with \textsc{Sampling} using $\chi^2$ yielding the lowest accuracy of 0.53. Similar trends were observed for SVM (square marker), Random Forest (diamond marker), and SGD (cross marker). A notable exception was Naive Bayes (triangle marker), where \textsc{Mascara} outperformed other approaches, achieving an accuracy of 0.73 compared to the baseline's 0.79. In general, Naive Bayes performs poorly due to its conditional independence assumption which does not align with preserving the correlation between individual features and the label while masking.

On the HCI dataset (Figure~\ref{fig:accuracy_d2_all}), we observed that all methods performed similarly well, except again when using Naive Bayes.

On the Income dataset (Figure~\ref{fig:accuracy_d3_all}), \framework using $g_3$ achieved an accuracy of 0.83, comparable to the baseline, for both Logistic Regression (circle marker) and SVM (square marker).
For Random Forest (diamond marker), \framework with MI and $\chi^2$ correlation measures yielded the best masked datasets, achieving an accuracy of 0.81 compared to the baseline's 0.83. When using SGD, both \framework and \textsc{Sampling} attained similar accuracies of 0.81, compared to the baseline's 0.85.

\begin{figure*}[t]
  \centering
  \begin{subfigure}{0.48\linewidth}
  \includegraphics[width=1\linewidth]{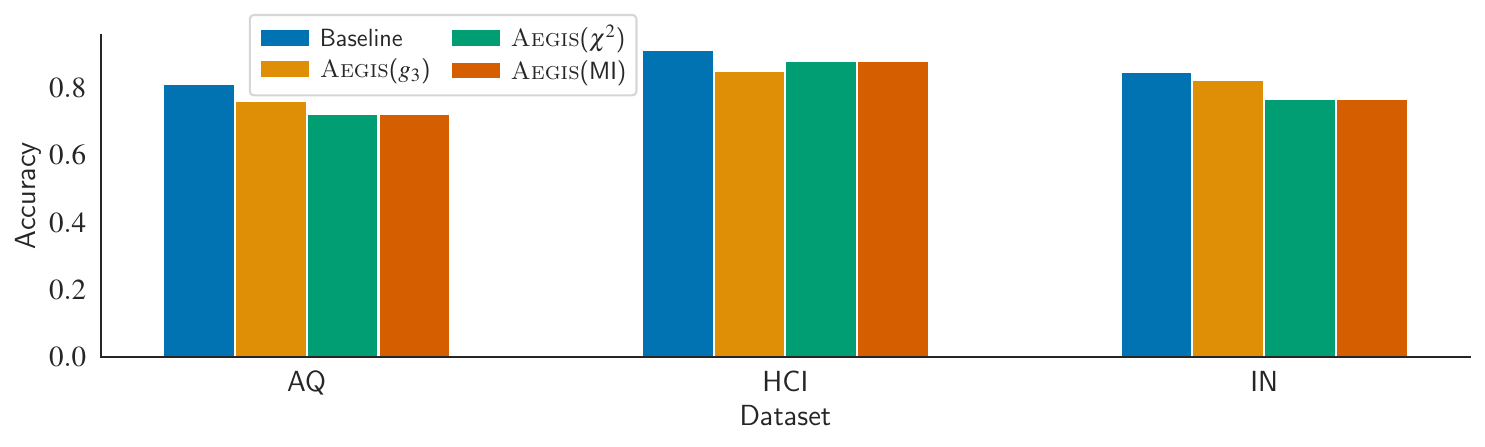}
  \caption{Effectiveness with Neural Network classifier}
  \label{fig:nn}
  \end{subfigure}\hfill
  \begin{subfigure}{0.48\linewidth}
    \centering
    \includegraphics[width=1\linewidth]{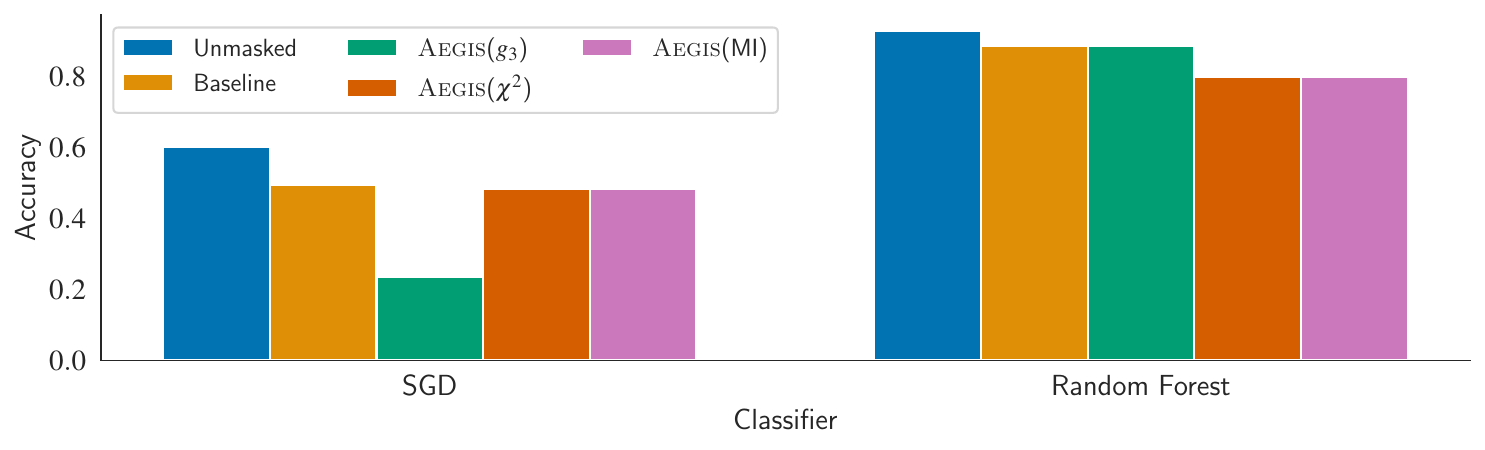}
  \caption{Effectiveness on $k$-anon.-based configurations.}
  \label{fig:k-anon}
  \end{subfigure}
  \caption{Effectiveness of \framework{} (a) with Neural Network classifier and (b) on $k$-anonymization-based configurations.}
\end{figure*}

Additionally, we compared the effectiveness of \framework with a Neural Network classifier. We use a fully connected 4-layer MLP classifier with two hidden layers of 100 neurons each and ReLU activation, trained using the Adam optimizer for up to 200 epochs. The network processes data through one-hot encoding for categorical features and evaluates performance using a 70-30 train-test split. The results are shown in Figure~\ref{fig:nn}. We observe that \framework (with $g_3$) achieves comparable performance to baseline: the accuracy was 6\% and 2\% less for AQ and IN datasets, resp.; and was 3\% less for HCI dataset when using with $\chi^2$ and MI.

We note that, although, out initial experiments with Neural networks are encouraging, they require more extensive evaluation due to their sensitivity to architectural choices, hyperparameters, and training dynamics, which can significantly impact performance. We leave this as an important future work.

\emph{Overall, \framework offers a model-agnostic framework to effectively determine a masking configuration that preserves key feature-label correlations required for high predictive utility, without ever training the downstream models during configuration selection.}

\subsubsection{Effectiveness on $k$-anonymization-based configurations}
\label{ssub:effectiveness_on_k_anonymization_based_configurations}

In these set of experiments, we studied the effectiveness of \framework when using real-world masking configurations based on $k$-anonymization (setting $k=5)$. We used the AQ dataset and two classifiers: Stochastic Gradient Decent (SGD), and Random Forest (RF). Our goal was to evaluate the accuracy of the masked AQ dataset determined by \framework{} and compare it the masked dataset determined by the baseline approach. As reference, we also evaluate the accuracy of unmasked dataset. The results are shown in Figure~\ref{fig:k-anon}. 

We observe that \framework achieves comparable performance for the two classifiers: the accuracy was 2\% less for SGD (with $\chi^2$) and was at par for RF (with $g_3$). These results are consistent with our observations when using synthetic masking configurations above as \framework, given a set of masking configurations as input, aims to determine the one that best preserves feature label correlations for a given correlation measure.

\subsection{Quality of Joint Distribution Reconstruction} % (fold)
\label{sub:accuracy_of_ipf_based_reconstruction}

In these experiments, we evaluate the quality of the reconstructed joint distributions produced by \framework+1D and \framework-1D. In both scenarios, we measure reconstruction fidelity using \textit{Total Variation Distance (TVD)}, a standard divergence metric that quantifies the difference between two probability distributions. Formally, given two discrete distributions \( P \) and \( Q \) over the same domain \( \Omega \), TVD is defined as:
\[
\mathrm{TVD}(P, Q) = \frac{1}{2} \sum_{\omega \in \Omega} \left| P(\omega) - Q(\omega) \right|
\]
This value represents the maximum shift in probability mass required to transform one distribution into the other. Lower TVD values indicate a closer match between the reconstructed and ground truth distributions.

\begin{figure}[t]
    \centering
    \includegraphics[width=0.7\linewidth]{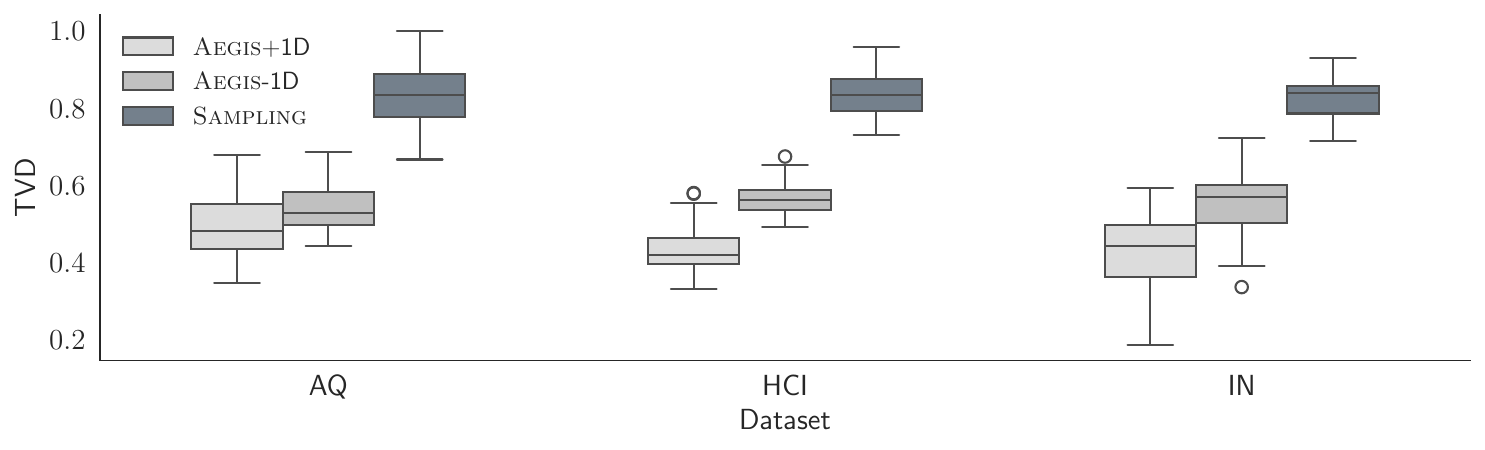}
    \caption{Quality of joint distribution reconstruction across 50 masking configurations.}
    \label{fig:ipf_quality}
\end{figure}

The results are shown in Figure~\ref{fig:ipf_quality}, where for each of the three datasets, we report the box plots over 50 different configurations (recall Algorithm~\ref{alg:recon}) for both \framework variants and for the sampling-based reconstruction. In every case, \framework{} achieves substantially lower TVD; the median TVD for \framework{}+1D lies between 0.45 and 0.50 across datasets, and even for \framework-1D, the median remains below 0.55. In contrast, the sampling-based reconstruction exhibits median errors of 0.80-0.85 and frequently reaches TVD values near 1.0, indicating almost complete distortion of the joint distribution. 

\emph{Overall, these results demonstrate that \framework's reconstruction using Iterative Proportional Fitting and modeling constraints derived from masking functions and data summaries, when available, yields more accurate reconstructed joint distributions than the sampling-based approach. In addition, this underscores \framework{}'s capability in producing model-agnostic utility estimates based on joint (feature-label) distributions for several downstream tasks in data-sharing ecosystems.}

\begin{figure*}[t]
    \centering
\begin{subfigure}{0.48\linewidth}
    \centering
    \includegraphics[width=\columnwidth]{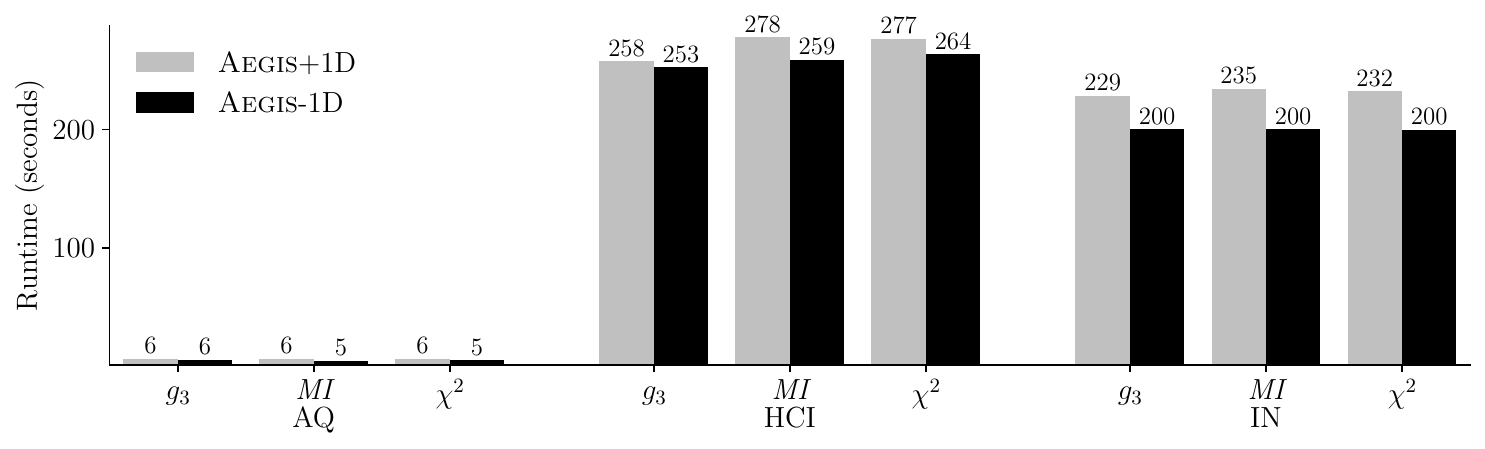}
\caption{Runtime}
\label{fig:impact_of_marginals_on_runtime_}
\end{subfigure}  \hfill
    \begin{subfigure}{0.48\linewidth}
    \centering
    \includegraphics[width=\columnwidth]{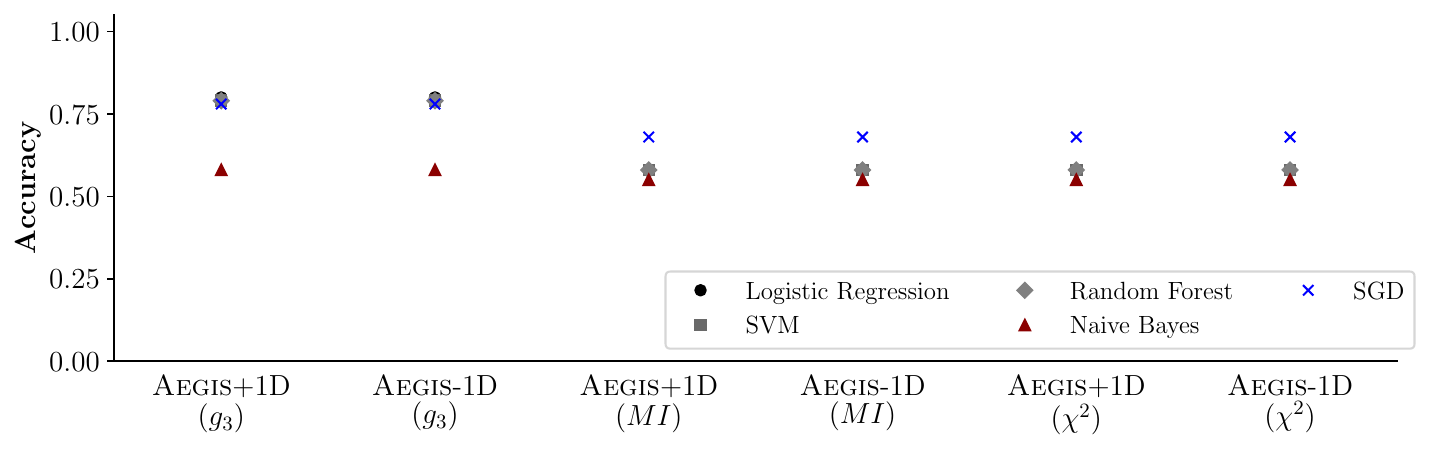}
    \caption{Accuracy on AQ dataset}
    \label{fig:accuracy_d1_1d_vs_nod}
  \end{subfigure}%

% \vspace{0.5em} % spacing between subfigures

\begin{subfigure}{0.48\linewidth}
    \centering
    \includegraphics[width=\columnwidth]{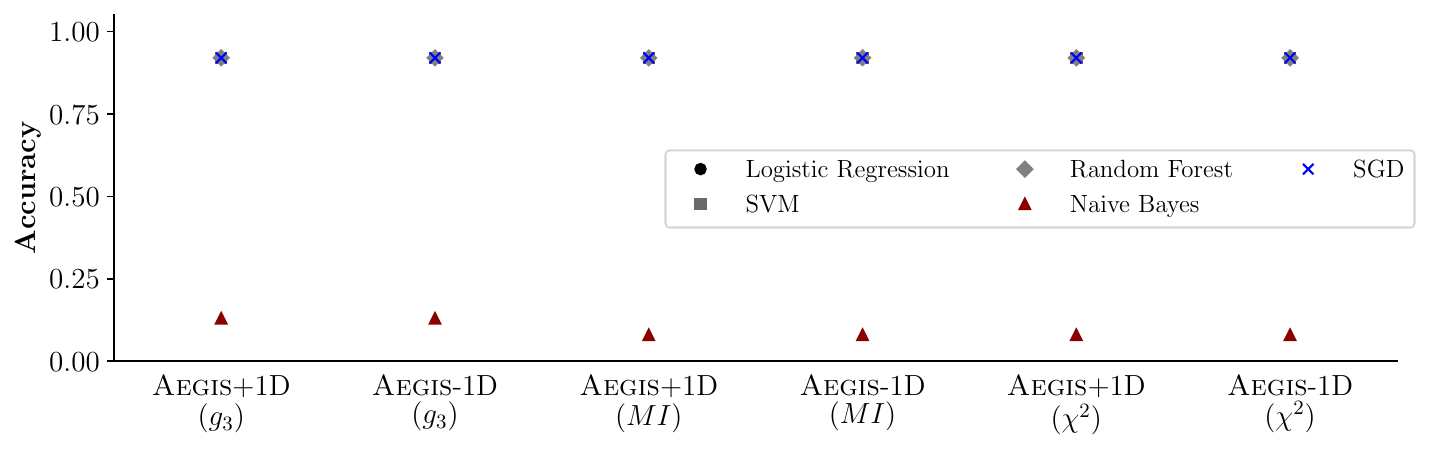}
    \caption{Accuracy on HCI dataset}
    \label{fig:accuracy_d2_1d_vs_nod}
  \end{subfigure}\hfill
\begin{subfigure}{0.48\linewidth}
    \centering
    \includegraphics[width=\columnwidth]{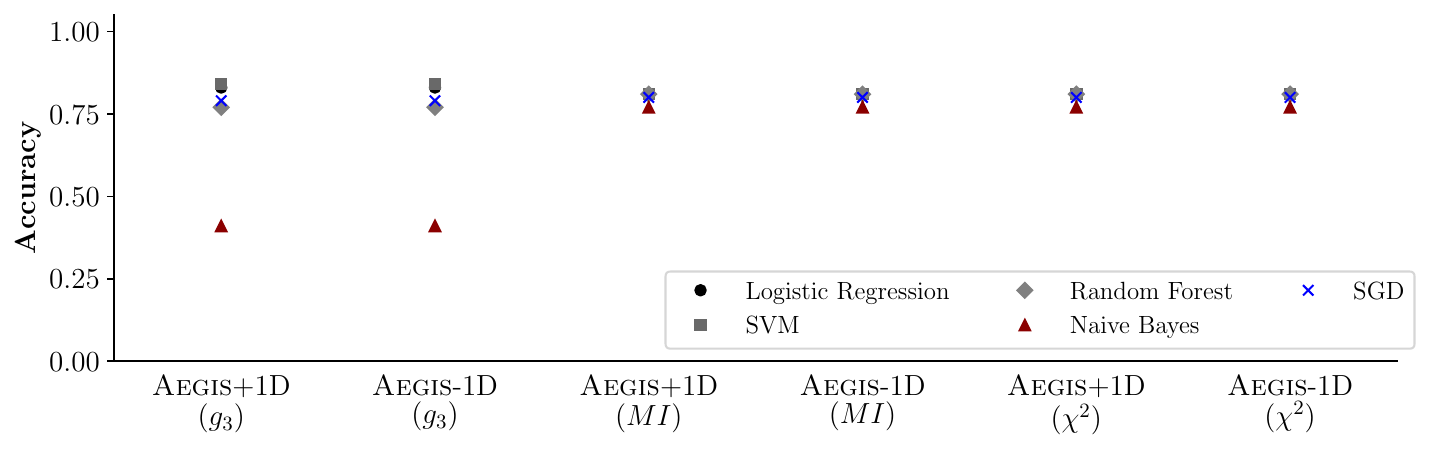}
    \caption{Accuracy on IN dataset}
    \label{fig:accuracy_d3_1d_vs_nod}
  \end{subfigure}

  \caption{Impact of availability of data summaries on accuracy of downstream tasks.}
  \label{fig:impact_of_marginals}
\end{figure*}

\subsection{Impact of Availability of Data Summaries} % (fold)
\label{sub:impact_of_different_data_summaries}

We now study how the availability of statistical summaries affects the performance in terms of runtime and downstream predictive utility (accuracy). The results are shown in Figure~\ref{fig:impact_of_marginals}, where we compare \framework{}+1D and \framework-1D. 

We first discuss the runtime experiments (see Figure~\ref{fig:impact_of_marginals_on_runtime_}) for \framework{} across the three model-agnostic measures and the three datasets. We observe that on the smaller AQ dataset, both variants complete in under $6.5s$. As the dataset size grows, the runtimes increase to roughly $230-280s$, where \framework{}-1D has a slightly better (up to 15\%) efficiency. This is expected as IPF converges faster with fewer constraints to satisfy. 

Figures~\ref{fig:accuracy_d1_1d_vs_nod}-\ref{fig:accuracy_d3_1d_vs_nod} show that the downstream predictive accuracy of four off-the-shelf classifiers (logistic regression, SVM, random forest, naive Bayes, and SGD) when trained on the data masked according to the configuration selected by \framework{}+1D and \framework{}-1D, for each of the three measures. In each case, \framework{}+1D yields higher accuracy than \framework{}-1D. For example, on the AQ dataset (Figure~\ref{fig:accuracy_d1_1d_vs_nod}), SVM accuracy improves by up to 2\%. We also observe similar improvements for HCI and IN datasets (Figures~\ref{fig:accuracy_d2_1d_vs_nod} and ~\ref{fig:accuracy_d3_1d_vs_nod}) across all measures. Naive Bayes, which is most sensitive to distributional errors, benefits most from the availability of data summaries.

\emph{Overall, we conclude that incorporating data summaries adds negligible overhead and helps determine better masking configurations. Additionally, even without data summaries, \framework{} still provides a robust framework to determine effective masking configurations.}

\begin{figure*}
\begin{subfigure}{.48\linewidth}
    \centering
    \includegraphics[width=\linewidth]{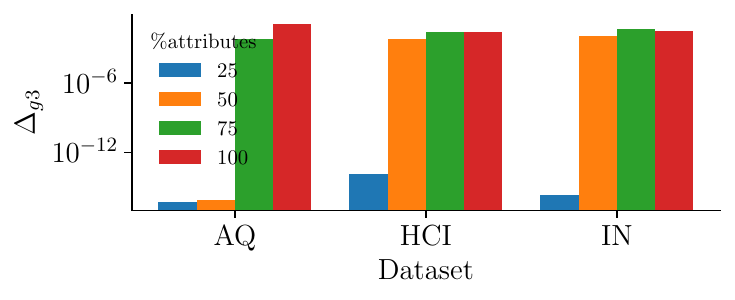}
    \caption{Impact on $\Delta_{g3}$}
    \label{fig:attribute_g3}
  \end{subfigure}
  \begin{subfigure}{.48\linewidth}
    \centering
    \includegraphics[width=\linewidth]{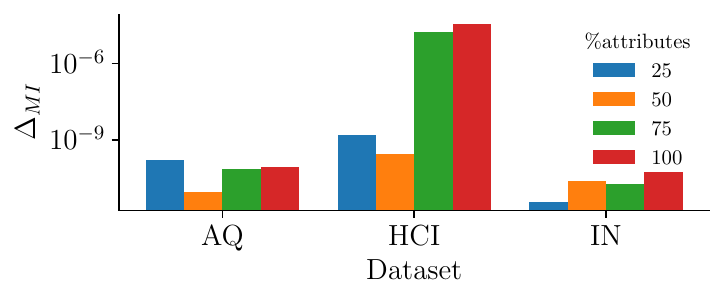}
    \caption{Impact on $\Delta_{MI}$}
    \label{fig:attribute_mi}
  \end{subfigure}
    \begin{subfigure}{.48\linewidth}
    \centering
    \includegraphics[width=\linewidth]{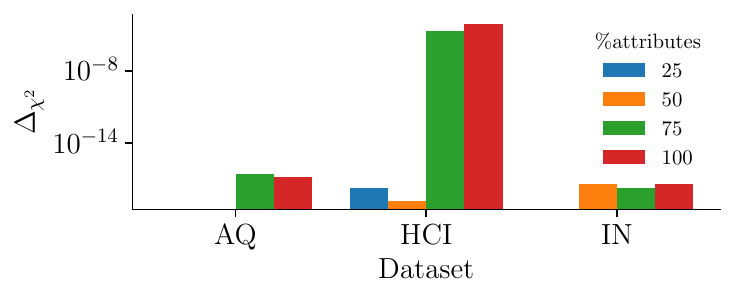}
    \caption{Impact on $\Delta_{\chi^2}$}
    \label{fig:attribute_chi2}
  \end{subfigure}
  \begin{subfigure}{.48\linewidth}
    \centering
    \includegraphics[width=\linewidth]{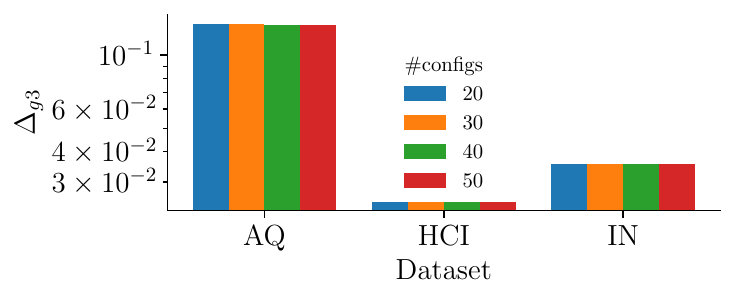}
    \caption{Impact on $\Delta_{g3}$}
    \label{fig:config_g3}
  \end{subfigure}
  \begin{subfigure}{.48\linewidth}
    \centering
    \includegraphics[width=\linewidth]{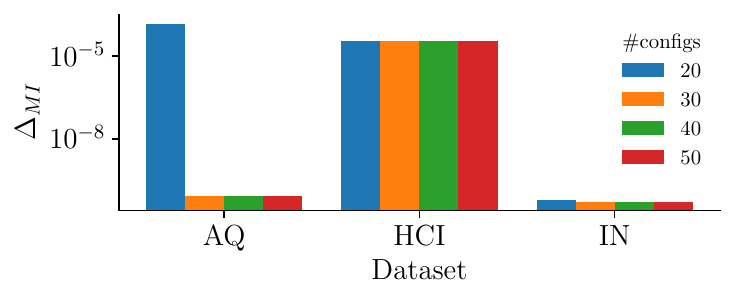}
    \caption{Impact on $\Delta_{MI}$}
    \label{fig:config_mi}
  \end{subfigure}
    \begin{subfigure}{.48\linewidth}
    \centering
    \includegraphics[width=\linewidth]{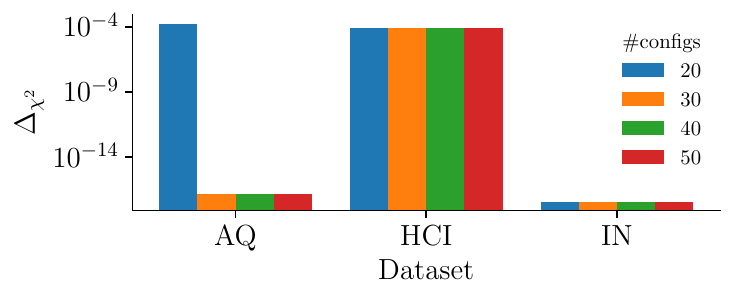}
    \caption{Impact on $\Delta_{\chi^2}$}
    \label{fig:config_chi2}
  \end{subfigure}
  \begin{subfigure}{.48\linewidth}
    \centering
    \includegraphics[width=\linewidth]{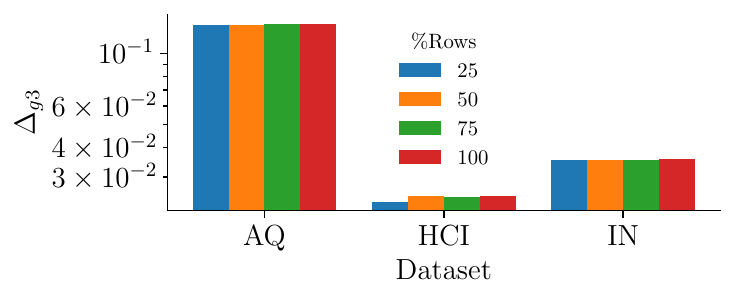}
    \caption{Impact on $\Delta_{g3}$}
    \label{fig:rows_g3}
  \end{subfigure}
  \begin{subfigure}{.48\linewidth}
    \centering
    \includegraphics[width=\linewidth]{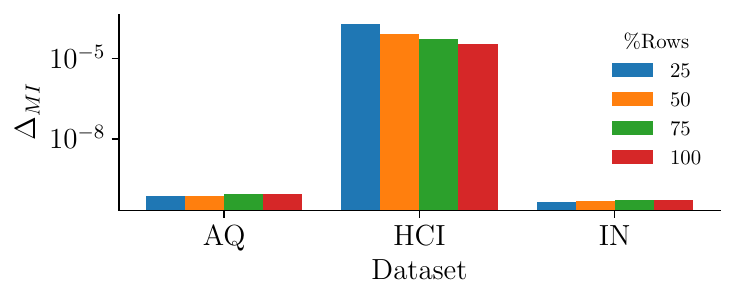}
    \caption{Impact on $\Delta_{MI}$}
    \label{fig:rows_mi}
  \end{subfigure}
    \begin{subfigure}{.48\linewidth}
    \centering
    \includegraphics[width=\linewidth]{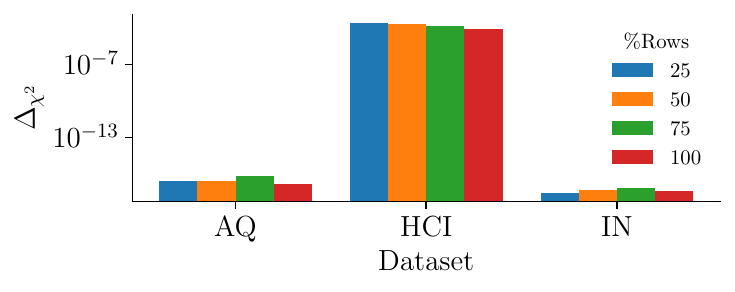}
    \caption{Impact on $\Delta_{\chi^2}$}
    \label{fig:rows_chi2}
  \end{subfigure}

\caption{Impact of parameters (\#~attributes (a)-(c), \#~configurations (d)-(e), and \#~rows (g)-(i)) on model-agnostic measures.}
\label{fig:impact_of_parameters}    
\end{figure*}

\subsection{Impact of Parameters} % (fold)
\label{sub:impact_of_parameters}

In this set of experiments, we evaluate the robustness of \framework to changes in three key parameters: the number of attributes (Figures~\ref{fig:attribute_g3}-\ref{fig:attribute_chi2}), the number of masking configurations (Figures~\ref{fig:config_g3}-\ref{fig:config_chi2}), and the number of rows in the dataset (Figures~\ref{fig:rows_g3}-\ref{fig:rows_chi2}). We measure the impact of these parameters on the three model-agnostic measures $g_3$, mutual information (MI), and Chi-squared statistic ($\chi^2$). The $\Delta$ on the y-axis (note the log scale) captures the change in predictive utility due to changes in these parameters.

\subsubsection{Impact of Number of Attributes}
We varied the number of attributes from 25\% to 100\% of the dataset's features. The results are shown in Figures~\ref{fig:attribute_g3}-\ref{fig:attribute_chi2}. Across all datasets and all three measures, we observe that the change in utility scores ($\Delta_{g_3}$, $\Delta_{MI}$, and $\Delta_{\chi^2}$) remains consistently small (below $10^{-8}$ for $\Delta_{g_3}$, $10^{-6}$ for $\Delta_{MI}$, and even lower for $\Delta_{\chi^2}$). The results show that \framework remains stable in computing the predictive utility when the dimensionality of the data varies significantly.

\subsubsection{Impact of Number of Configurations}
Figures~\ref{fig:config_g3}-\ref{fig:config_chi2} show the effect of changing the number of masking configurations from 20 to 50. We observe negligible differences in the utility scores across all three datasets. For example, $\Delta_{\chi^2}$ remains below $10^{-7}$, and $\Delta_{g_3}$ and $\Delta_{MI}$ show only minor fluctuations (on the order of $10^{-2}-10^{-6}$). The experiments demonstrate that \framework{} is not sensitive to the exact number of candidate masking configurations.

\subsubsection{Impact of Number of Rows}
Lastly, we evaluate the effect of increasing the sample size from 25\% to 100\% of the dataset. The results are shown in Figures~\ref{fig:rows_g3}-\ref{fig:rows_chi2}. We observe that the changes in the utility metrics are negligible, particularly for $\Delta_{MI}$ and $\Delta_{\chi^2}$, where the differences are on the order of $10^{-10}$ or less. Even for $\Delta_{g3}$, which shows the largest variation, the overall change remains small and bounded.

\emph{In sum, the impact of varying attributes, configurations, and rows is minimal, both in absolute terms and relative to the log. y-axis scale. The largest observed change ($\Delta_{g3}$ in the AQ dataset) remains below $10^{-1}$, while most other variations are several orders of magnitude smaller. This demonstrates the \framework's robustness and stability.}

\subsection{Impact of Generalization and Domain Size}
\label{sec:impact_of_generalization_and_domain_size}

\begin{figure*}[t]
  \begin{subfigure}{0.4\linewidth}
    \centering
    \includegraphics[width=\linewidth]{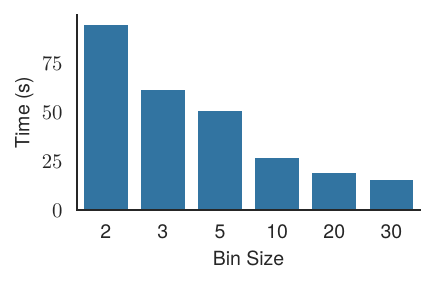}
    \caption{Reconst. Time vs. Bucket Size}
    \label{fig:time-bin}
  \end{subfigure}
  \begin{subfigure}{0.4\linewidth}
    \centering
    \includegraphics[width=\linewidth]{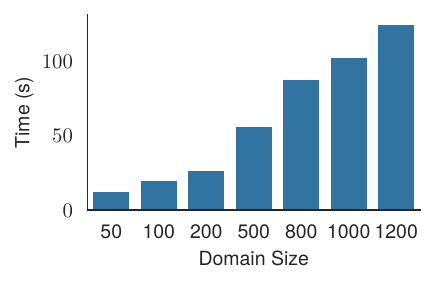}
    \caption{Reconst. Time vs. Domain Size}
    \label{fig:time-domain}
  \end{subfigure}

  \begin{subfigure}{0.4\linewidth}
    \centering
    \includegraphics[width=\linewidth]{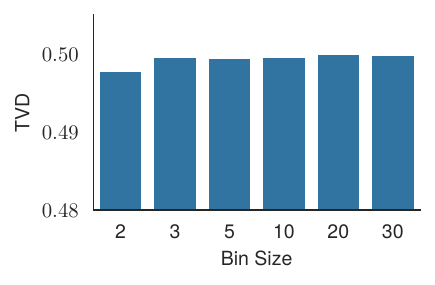}
    \caption{TVD vs. Bucket Size}
    \label{fig:tvd-bin}
  \end{subfigure}
    \begin{subfigure}{0.4\linewidth}
    \centering
    \includegraphics[width=\linewidth]{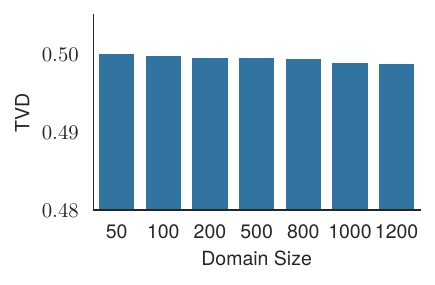}
    \caption{TVD vs. Domain Size}
    \label{fig:tvd-domain}
  \end{subfigure}
\caption{Impact of generalization and domain size on joint distribution reconstruction.}
\label{fig:impact-domain-bin}
\end{figure*}

In this set of experiments, we evaluate the impact of generalization and domain size on the robustness of \framework’s joint–distribution reconstruction. We construct a synthetic dataset comprising numerical feature drawn from a uniform distribution and a binary class label, with 10 000 records. To study generalization, in each run we generate configurations with increasing bucket sizes, thereby varying the degree of value coarsening. Additionally, we consider runs with varying the feature’s domain size by changing its number of distinct values. This design allows us to isolate and quantify how both generalization (via bucket size) and domain cardinality affect reconstruction quality. The results are shown in Figure~\ref{fig:impact-domain-bin}.

We first discuss the impact on reconstruction time. Figure~\ref{fig:time-bin} shows that reconstruction time decreases as the bucket (bin) size increases with domain fixed at 200. This is because when the feature domain is partitioned into very fine‐grained buckets (e.g., size = 2), IPF must reconcile a larger number of probability cells, leading to runtimes on the order of 90s. As the bucket size grows to 5 and then 10, the number of cells falls, and reconstruction time drops to approximately 50s and 25s, respectively. Beyond a bucket size of 20, the dimensionality of the contingency table is sufficiently low, and reconstruction requires much less time. Figure~\ref{fig:time-domain} shows the complementary effect of domain cardinality with bin size fixed at 10. With just 50 distinct values, reconstruction completes in under 15s; however, as we enlarge the domain to 500 values, time increases to 55s, and exceeds 120s when the domain reaches 1200 values. This nearly linear growth in runtime reflects the cost of performing IPF over ever‐larger joint‐distribution tables.

We also evaluated the impact on reconstruction quality using TVD (see above) in Figures~\ref{fig:tvd-bin} and \ref{fig:tvd-domain}. We observe that (TVD) increases slightly as the bin size grows—indicating a marginal degradation in reconstruction fidelity when the data are more coarsely generalized. Conversely, TVD decreases as the domain size expands, reflecting improved fidelity when more distinct values are available. These TVD mirror our observations on the AQ, HCI, and IN datasets (see Section~\ref{sub:accuracy_of_ipf_based_reconstruction}), where TVD values remain below 0.5.

\emph{Overall, increasing bin size reduces reconstruction runtime but slightly increases TVD, whereas increasing domain size raises runtime while marginally lowering TVD.}

\subsection{Framework Efficiency} % (fold)
\label{sub:framework_efficiency}

\begin{figure*}
    \begin{subfigure}{.48\linewidth}
    \centering
    \includegraphics[width=\linewidth]{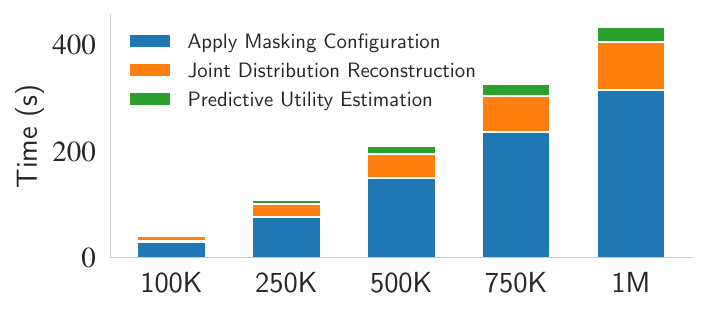}
    \caption{Dataset Size (\#Rows)}
    \label{fig:scale_rows}
  \end{subfigure}
      \begin{subfigure}{.48\linewidth}
    \centering
    \includegraphics[width=\linewidth]{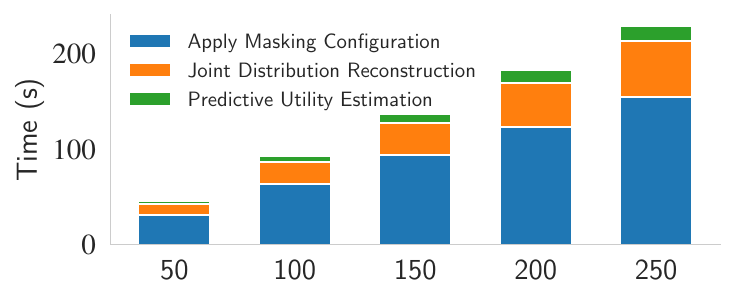}
    \caption{Number of Attributes}
    \label{fig:scale_columns}
  \end{subfigure}

    \begin{subfigure}[t]{.48\linewidth}
    \centering
    \includegraphics[width=\linewidth]{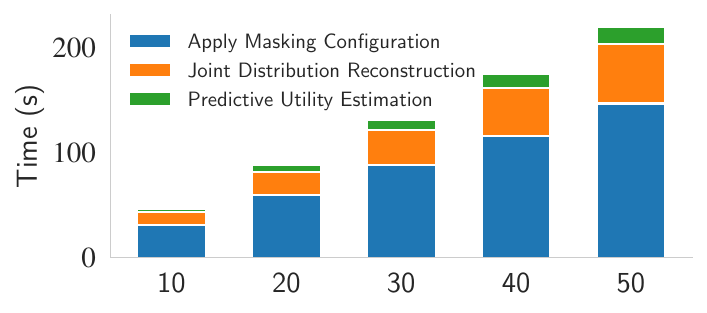}
    \caption{Number of Configurations}
    \label{fig:scale_configs}
  \end{subfigure}
  \begin{subfigure}[t]{0.48\linewidth}
  \centering
  \includegraphics[width=\linewidth]{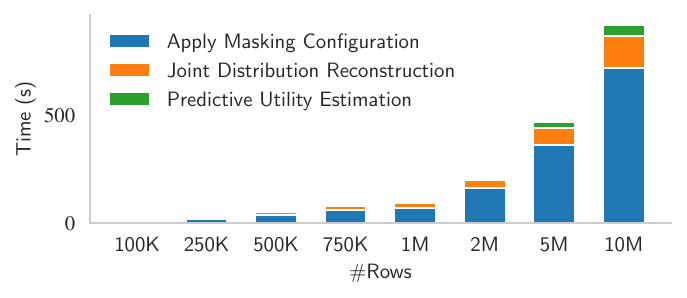}
  \caption{Additional scalability experiment w.r.t. \#rows.}
  \label{fig:scale_rows_add}
  \end{subfigure}
    
\caption{Scalability Experiments}
\label{fig:scalability_experiments}    
\end{figure*}

In our last set of experiments, we study the scalability of our approach by measuring the runtime with varying numbers of rows, attributes, and masking configurations. We used a synthetic dataset (as described earlier) for these experiments, for which the results are shown in Figures~\ref{fig:scale_rows}-\ref{fig:scale_configs}. Each figure shows the runtime breakdown for applying the masking configuration, reconstruction, and utility estimation.

As Figure~\ref{fig:scale_rows} shows, the runtime increases steadily with the number of rows (from 100K to 1M). We observe that applying masking configuration to compute the masked joint distribution dominates the runtime, with reconstruction and utility estimation adding less overhead. For instance, for 1M rows, the total runtime exceeds $400s$ with computing masked joint distribution accounting for more than 50\%. This linear growth is expected as more rows increase the masking operation and reconstruction routing (recall Section~\ref{sub:reconstructing_joint_distributions}).

In Figure~\ref{fig:scale_columns}, we vary the number of attributes from 50 to 250. We observe that runtime increases steadily. This is again because of the increasing cost of applying the masking configurations, which depends on the number of attributes. 

We also examine the scalability with respect to the number of masking configurations (from 10-50; Figure~\ref{fig:scale_configs}). We observe that runtime grows approximately linearly across this range, again dominated by the masking. The reconstruction cost increases linearly, although more slowly, while the utility metric evaluation remains consistently low (less than 10\%).

Lastly, we also evaluated scalability on larger datasets upto 10M rows. 
The results are shown in Figure~\ref{fig:scale_rows_add}. We note up front that these large-scale experiments were run on a more powerful hardware 
configuration (with GPU acceleration) than the CPU‐only setup described in Section~\ref{sub:experimental_setup}; on the CPU-only we could not scale beyond 1M rows in a reasonable time. From Figure~\ref{fig:scale_rows_add}, we see similar trends in that time grows roughly linearly with the number of records. Applying the masking configuration dominates the cost, rising from under 20s at 100K rows to roughly 740s at 10M rows. Joint distribution reconstruction remains in the the order of 15--20\%—and predictive utility estimation contributes only a few seconds (up to 53s) even at the largest scale.  Together, these results confirm that \framework can handle large datasets when GPU resources are available. For even larger dataset (with 1B rows or more), runtimes would clearly exceed practical budget for selecting configurations. As future work, we plan to parallelize algorithms within \framework{} to further reduce end-to-end runtime and enable fast configuration selection on truly massive datasets.

\emph{Overall, \framework{} demonstrates scalable performance across dataset size, dimensionality, and a number of masking configurations.}

%!TEX root=./main.tex

\section{Related Work} % (fold)
\label{sec:related_work}

We now discuss how ideas presented in this paper relate to prior works, which can be broadly categorized into:

\paragraph{Data Masking}

Data Masking is a well-studied topic in the privacy literature. Beyond modifying or suppressing identifiers, anonymization methods including $k$-Anonymity, $l$-Diversity, and $t$-Closeness~\cite{sweeney2002k,Machanavajjhala2006ldiversity,Li2007tcloseness} are based on the concept of equivalence classes and group records to satisfy privacy thresholds. The original data can be anonymized using these techniques, each distorting correlations to varying degrees. Differential privacy~\cite{dwork2006differential} based masking techniques offer strong privacy guarantees. However, noise often degrades feature‑label dependencies, leading to utility‑aware variants~\cite{li2009tradeoff,miklau2022negotiating}. Graph and relational anonymization techniques~\cite{10.14778/1453856.1453947,ghinita2007fast} have also been studied to protect link structure~\cite{Nergiz2007,terrovitis2012privacy}. In general, masking often involves masking functions like generalization, suppression, perturbation, tokenization, and shuffling to obscure raw values while aiming to retain aggregate statistics.  
Our work complements all of these masking techniques---\framework can incorporate any $k$-anonymity, differential privacy, structural anonymization, or masking operation to evaluate and select the configuration that best preserves predictive utility in a model-agnostic way.

\paragraph{Data Correlation}

Maintaining feature–label and inter‑feature dependencies is critical for downstream analytical tasks. Key measures include the Chi‑Square Test for categorical independence; Mutual Information for capturing linear and nonlinear associations; Functional‑Dependency Errors (G1, G2, G3)~\cite{KIVINEN1995129} for quantifying violations of $X\!\to\!Y$; Correlation Matrices and Partial Correlation, for multivariate and conditional relationships; KL/Jensen‑Shannon divergence measures, for distributional shifts. These metrics evaluate the effect of a given transformation but do not guide the \emph{selection} of transformations that minimize correlation distortion. \framework{} can incorporate model agnostic measures based on joint distributions, which allows selecting the masking configuration that best preserves predictive utility without ever training a downstream model. An interesting direction of future work is to extend our framework of other correlation measures as well.

\paragraph{Balancing Privacy and Utility}
Optimizing data utility under privacy requirement is a major challenge in using anonymized data for ML~\cite{li2022federated, cormode2013empirical, soria2014enhancing}. \cite{rastogi2007boundary} delves into the fundamental tradeoff between privacy and utility when publishing data for counting queries. \cite{ghinita2007fast} tackles the problem of anonymizing data with the least information loss before releasing it. \cite{terrovitis2012privacy} focuses on publishing sparse, multidimensional data, like web search logs. \cite{stadler2022synthetic} quantitatively evaluated the trade-off between privacy protection and data utility of synthetic data than traditional anonymization methods. \cite{miklau2022negotiating} advocates adopting differential privacy frameworks for publishing aggregated data. \cite{li2009tradeoff} have studied balancing privacy protection and data utility when publishing anonymized microdata. In the context of our setting, \framework{} is not itself a privacy mechanism and treats privacy as binary, i.e., given a set of privacy-guaranteed candidate configurations, it selects the one (without access to raw dataset and using available summaries or none) that maximizes feature–label correlation under the chosen correlation metric.

\paragraph{Data Reconstruction}
Reconstruction of distribution has also been studied in various context. For example, Mosaic~\cite{DBLP:conf/cidr/OrrAJCBS20} utilizes IPF to re-weight samples for answering population queries when the sampling mechanism is unknown. \cite{10.14778/3231751.3231752} address the problem of signal reconstruction in databases and leverage IPF. When initial solutions yield negative values. \cite{10.14778/1687627.1687642} employs IPF within an entropy-maximization framework to estimate co-occurrence probabilities for keyword-query expansions. \cite{10.14778/2556549.2556560} presents a framework to address the empty-answer problem in databases by estimating joint distributions from marginal data, facilitating the generation of probable answers when exact matches are absent. In our work, we leverage IPF to reconstruct joint distributions for computing predictive utility deviation. The key difference lies in modeling constraints that are specific to our setting.

\paragraph{Data-sharing Ecosystems}
Finally, our work also intersects with research on data‑sharing ecosystems, e.g. ~\cite{10.1145/3456859.3456861, gaia-x, dawex, aws, snowflake, openmined2023, datashield2023, iudx, gaiax2023, fainder-vldb, fainder-demo-sigmod}. Whereas those efforts predominantly explore the underlying infrastructure, system architectures, and policy trade‑offs that enable secure, scalable data exchange, \framework offers a complementary, task‑agnostic framework that allows data publishers and providers to share anonymized datasets without sacrificing predictive utility.

\section{Conclusion} \label{sec:conclusion}

We introduced \framework{}, a middleware framework designed to identify optimal masking configurations for machine learning datasets to preserve their predictive utility without requiring the ML pipeline execution. In data-sharing ecosystems where raw data is anonymized and inaccessible, \framework{} serves as an ``advisor'' to data providers, recommending masking strategies that maximize utility in a task-agnostic manner. The framework builds on the understanding that predictive utility could be estimated using model-agnostic measures like correlation or association. \framework{} is compatible with any utility estimator that relies on joint distribution estimation and is instantiated with well-established techniques like Mutual Information, Chi-Square, and functional dependency-based measures. We studied computational challenges in designing \framework{}. A central computational challenge in \framework{} lies in estimating joint distributions. We formalized this as a constrained optimization problem and adapted Iterative Proportional Fitting (IPF) to solve it efficiently. We evaluated \framework{} on real-world datasets and downstream models, demonstrating that it achieves comparable predictive accuracy to exhaustive pipeline while offering significant speed-ups—often by order of magnitude. Our experiments also showed that different algorithms within \framework{} are highly effective and scalable. 

As ongoing work, we explore how \framework{} can integrate correlation measures beyond those based on attribute-label joint distributions. In the future, we aim to design masking functions that jointly preserve predictive utility and privacy, explore how masking strategies interact with Neural Networks, LLMs or foundational models when applied to tabular data, and parallelize \framework{} to scale to massive datasets.

\section*{Acknowledgments}

This work of Kaustubh Beedkar is supported by Anusandhan National Research Foundation project ANRF/ECRG/2024/005781/ENS. The work of Fatemeh Ramezani Khozestani, Sohrab Namazi Nia, and Senjuti Basu Roy are supported by the following funding agencies: (1) National Science Foundation award number(s): 1942913, 1814595 (2) Office of Naval Research award number(s): N000141812838, N000142112966, N000142412466.

\bibliographystyle{ACM-Reference-Format}
\bibliography{acmart}

\end{document}